\documentclass[runningheads]{llncs}
\pdfoutput=1
\usepackage[mobile]{eccv}

\usepackage{eccvabbrv}

\usepackage{graphicx}
\usepackage{booktabs}

\definecolor{cvprblue}{rgb}{0.21,0.49,0.74}
\usepackage{tcolorbox}

\usepackage{xcolor}  
\definecolor{mydarkblue}{rgb}{0.68, 0.85, 1.0}
\usepackage{xcolor}   
\usepackage[utf8]{inputenc} % allow utf-8 input
\usepackage[T1]{fontenc}    % use 8-bit T1 fonts
\usepackage{url}            % simple URL typesetting
\usepackage{booktabs}       % professional-quality tables
\usepackage{amsfonts}       % blackboard math symbols
\usepackage{nicefrac}       % compact symbols for 1/2, etc.
\usepackage{microtype}      % microtypography
\usepackage{sidecap}
\usepackage{float}
\usepackage{graphicx}

\usepackage{afterpage}
\usepackage{multirow}
\usepackage{mathrsfs}
\newcommand{\vhead}[1]{\rotatebox{60}{\parbox[c]{0cm}{\centering{#1}}}}
\usepackage{amsmath}
\usepackage{amssymb}
\usepackage{mathtools}
\usepackage[mathscr]{eucal}% 
\usepackage{bm}% bold math
\usepackage{bbm}
\usepackage{relsize}
\usepackage{graphics}
\usepackage{wrapfig}
\usepackage{multirow}
\usepackage{enumitem}
\usepackage{tablefootnote}
\usepackage{pifont} % for cmark
\usepackage{array}
\usepackage{color, colortbl}
\definecolor{Gray}{gray}{0.9}
\definecolor{lightskyblue}{rgb}{0.53, 0.81, 0.98}

\definecolor{softblue}{rgb}{0.85, 0.91, 0.98}
\definecolor{mygray}{gray}{0.85}
\definecolor{lightblue}{RGB}{27,161,226}
\definecolor{bg_lightblue}{RGB}{218,232,252}
\definecolor{df_purple}{RGB}{128,0,128}
\usepackage{algorithm}
\usepackage{algorithmic}
\usepackage{pgf}
\usepackage{xinttools}
\usepackage{tikz}
\usetikzlibrary{arrows.meta}
\usepackage{textcomp}
\usetikzlibrary{calc,shapes,arrows,positioning,automata,trees}
\usepackage{placeins} 
\usepackage{tabularx}
\usepackage{graphicx}
\usepackage{listings}

\def\eg{\emph{e.g.}}

\usepackage[font=small,labelfont=bf,tableposition=top]{caption}

\DeclareCaptionLabelFormat{andtable}{#1~#2  \&  \tablename~\thetable}

\usepackage{blindtext}

\newcommand{\cmark}{\ding{51}}
\newcommand{\xmark}{\ding{55}}

\newlength\savewidth

\makeatletter\renewcommand\paragraph{\@startsection{paragraph}{4}{\z@}
  {.5em \@plus1ex \@minus.2ex}{-.5em}{\normalfont\normalsize\bfseries}}\makeatother

\usepackage{bigdelim}
\usepackage{colortbl} % colo\newcommand{\soft}{\mathrm{softmax}}red cells
\definecolor{mColor1}{rgb}{0.95,0.95,0.95}

\usepackage{float}

\definecolor{tabhighlight}{HTML}{e5e5e5}

\usepackage{microtype}
\usepackage{graphicx}
\usepackage{booktabs} % for professional tables

\usepackage[accsupp]{axessibility}  % Improves PDF readability for those with disabilities.

\usepackage{orcidlink}

\begin{document}

\title{Training-Free Semantic Segmentation via LLM-Supervision}

\titlerunning{Training-Free Semantic Segmentation}

\author{Wenfang Sun\inst{1} \thanks{Equal contribution.}\and
Yingjun Du\inst{2,3}\thanks{Equal contribution. Work done during an internship at Cisco.} \and 
Gaowen Liu\inst{3}  \and \\
Ramana Kompella\inst{3}  \and
Cees G.M. Snoek\inst{2}}  
\authorrunning{Sun. et al.}
\institute{University of Science and Technology of China \and
AIM Lab, University of Amsterdam
\\
 \and
Cisco Research
\\
}

\maketitle

\begin{abstract}
Recent advancements in open vocabulary models, like CLIP, have notably advanced zero-shot classification and segmentation by utilizing natural language for class-specific embeddings. However, most research has focused on improving model accuracy through prompt engineering, prompt learning, or fine-tuning with limited labeled data, thereby overlooking the importance of refining the class descriptors.
This paper introduces a new approach to text-supervised semantic segmentation using supervision by a large language model (LLM) that does not require extra training. Our method starts from an LLM, like GPT-3, to generate a detailed set of subclasses for more accurate class representation. We then employ an advanced text-supervised semantic segmentation model to apply the generated subclasses as target labels, resulting in diverse segmentation results tailored to each subclass's unique characteristics.
Additionally, we propose an assembly that merges the segmentation maps from the various subclass descriptors to ensure a more comprehensive representation of the different aspects in the test images. 
Through comprehensive experiments on three standard benchmarks, our method outperforms traditional text-supervised semantic segmentation methods by a marked margin.
\end{abstract}    
\section{Introduction}
\label{sec:intro}

\begin{figure}[t]
    \centering
    \includegraphics[width=0.95\textwidth]{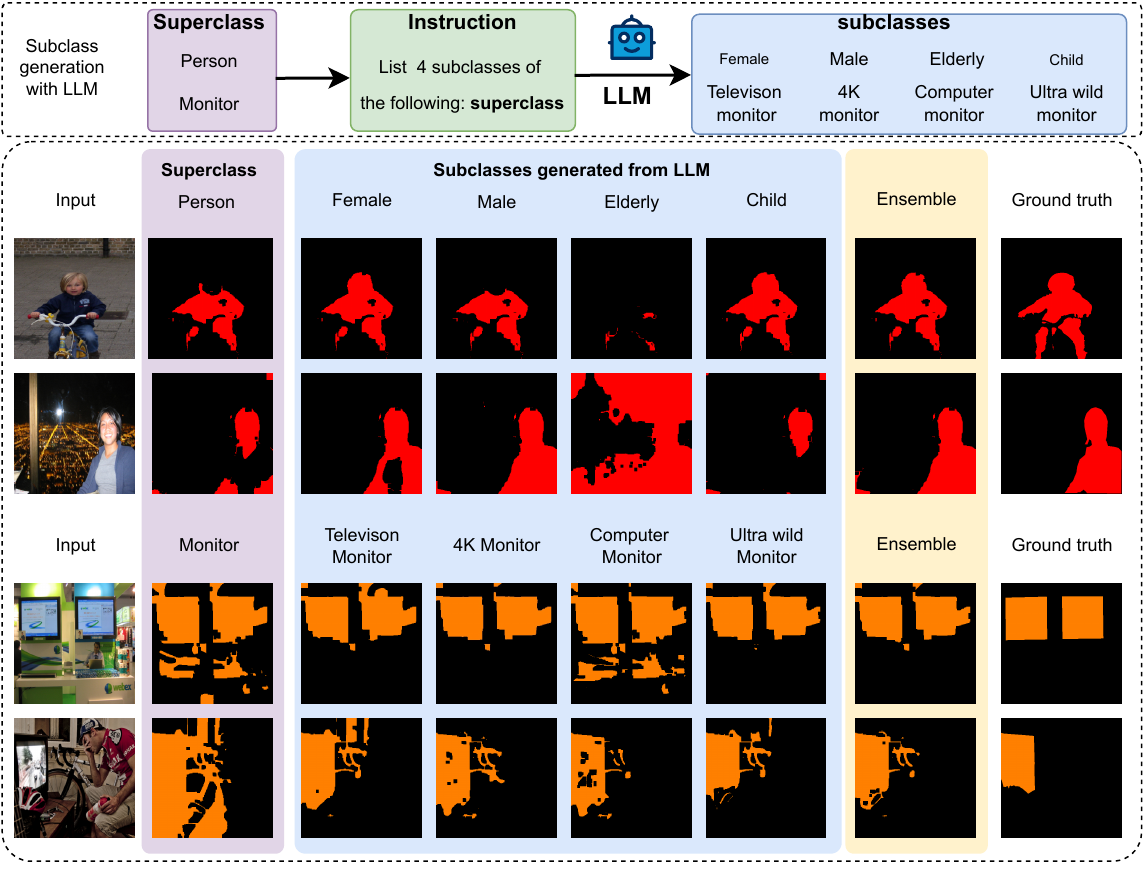}
\caption{
%\textbf{Subclass generation using LLM (top) and our training-free semantic segmentation results under LLM-supervision.}  
\textbf{Overview of the proposed approach.}
Initially, we start from an LLM to generate subclasses for a given superclass in the test phase. Subsequently, we utilize a pre-existing semantic segmentation model~\cite{SimSeg}, initially supervised by superclasses, and apply the LLM-generated subclasses for supervision. To obtain the final results, we ensemble the results from various subclasses(\textcolor{blue}{\textbf{blue}} areas).
Compared with the traditional superclass-supervised method (\textcolor{df_purple}{\textbf{purple}} area), our LLM-supervision  (\textcolor{yellow}{\textbf{yellow}} areas) achieves distinguishable and precise segmentation results. }
    \label{fig:intro_seg}
\end{figure}
Semantic segmentation plays a crucial role in computer vision, aiming to assign specific semantic classes to their respective pixels. The recent emergence of vision and language models~\cite{CLIP, li2022supervision, li2022blip, yao2021filip}, as a standard for generalized zero-shot learning, has given rise to a new area of research: language-driven semantic  segmentation~\cite{SimSeg, li2022languagedriven, ghiasi2022scaling}, which enables the creation of semantic segments through contrasting image and text. Recently, some studies~\cite{extractclip, xu2022simple, liu2022open, kirillov2023segment} have investigated segmentation based on CLIP to enhance transferability, focusing on areas such as zero-shot~\cite{xian2019semantic,bucher2019zero} and open-vocabulary segmentation~\cite{zhao2017open}. While recent advancements have primarily concentrated on improving model accuracy through techniques like prompt engineering~\cite{luddecke2022image}, prompt learning~\cite{liu2022prompt}, and selective fine-tuning with labeled data~\cite{wang2023iterative}, an essential yet less explored area is the refinement of the class descriptors themselves. These descriptors are critical to the accuracy and effectiveness of language-driven semantic segmentation models and the focus of this paper.

We are inspired by recent work on training-free vision language models, \eg,~\cite{CuPL, CHILS, mirza2023lafter, Waffle, oikarinen2023label}. These models utilize large language models (LLM)~\cite{GPT3} to generate additional descriptions per class, such as subclass names~\cite{CHILS}, class descriptions~\cite{Waffle, mirza2023lafter, CuPL}, or concept descriptions~\cite{oikarinen2023label}.
These approaches are based on generating additional descriptions to provide each class with more detailed and informative representations, specifically for the image classification problem. 
However, their application in semantic segmentation tasks has not yet been investigated, which is more challenging due to the need for precise pixel-level understanding and differentiation of various image elements.
In this paper, we propose a novel training-free semantic segmentation via LLM-supervision. 

We make three contributions, illustrated in Figure~\ref{fig:intro_seg}.
First, we suggest employing LLMs to generate subclass names for each class to tackle the problem of similar semantic features in categories found in traditional text-guided semantic segmentation techniques. This strategy aims to enhance the differentiation of features within the subclass space, distinguishing the original classes more clearly.
Second, we implement an advanced text-supervised semantic segmentation method~\cite{SimSeg}, using the generated subclass descriptors as the target labels. This approach facilitates the production of a wide variety of segmentation results. Notably, our model is training-free, eliminating the need for additional training in the semantic segmentation process.
Third, we introduce the ensembling of subclass descriptors, efficiently combining segmentation maps from different subclass descriptors with the original superclass representation. This integration is essential for conducting a more exhaustive and comprehensive analysis of test images, allowing for exploring a wider range of visual features than what is achievable with conventional methods.

Our method is characterized by its adaptability, easy integration, and compatibility with existing models, making it a straightforward yet effective enhancement to current text-supervised semantic segmentation frameworks. We demonstrate that our method outperforms conventional text-supervised segmentation methods through extensive empirical evaluation across three commonly used standard benchmarks. This improved performance underscores the efficacy of our approach in addressing the limitations of current models and paving the way for more accurate and efficient semantic segmentation.
\section{Related work}
\label{sec:related}
\textbf{Vision-Language Foundation Models.} Vision-Language Foundation Models represent an emerging and pivotal area of research that explores the intricate interplay between visual and language information. Within this domain, models like CLIP~\cite{CLIP}, FILIP~\cite{yao2021filip}, and ALIGN~\cite{ALIGN} stand out as prominent examples. They leverage vast web-curated image-text pairs to achieve impressive zero-shot transfer ability. They demonstrate their efficacy in learning robust image and text representations for cross-modal alignment and zero-shot image classification tasks. 
Vision-Language Foundation Models, with CLIP as a shining exemplar, showcase their adaptability across a broad spectrum of downstream tasks, including text-driven image manipulation~\cite{styleclip}, few-shot image recognition \cite{gao2021clip, zhang2021tip, kim2022how}, visual grounding~\cite{Cpt}, image captioning~\cite{Clipscore}, object detection~\cite{detclip,du2022learning,rasheed2022bridging}, and semantic segmentation~\cite{extractclip, Denseclip, SimSeg,ding2022decoupling}. In doing so, they effectively bridge the gap between computer vision and natural language processing.
Conversely, our approach introduces a training-free, text-supervised visual-language model for semantic segmentation. This model does not require extra training, reducing the computational resources needed.

\noindent\textbf{Label-free models.} Some previous works~\cite{CHILS, CuPL, Waffle, VisDesc,yan2023learning, oikarinen2023label,mirza2023lafter} have introduced label-free or training-free models, leveraging the robust and universal semantic representation capabilities of LLM~\cite{GPT3}. 
Label-free CBM~\cite{oikarinen2023label} utilizes an LLM to generate concept data without relying on explicit labels. This approach demonstrates scalable interpretability on ImageNet~\cite{deng2009imagenet}, maintaining high accuracy while minimizing the need for human involvement.
LaFTer~\cite{mirza2023lafter} harnesses the power of an LLM to bridge the performance gap in zero-shot visual classification. It significantly improves without requiring explicit labels or paired vision and language data.
CHiLS~\cite{CHILS} leverages hierarchical label sets generated by LLM to enhance the zero-shot image classification capabilities of the CLIP model. It achieves state-of-the-art performance in zero-shot classification datasets with semantic hierarchies.
VisDesc~\cite{VisDesc} introduces a method using detailed textual descriptors from LLM to boost image classification accuracy and interpretability. This is achieved by assigning weights to each descriptor and assessing their similarity to individual image attributes. 
CuPL~\cite{CuPL} introduces a novel approach that utilizes the capabilities of LLM to generate semantically rich texts, prompting CLIP and enhancing text-image alignment. This improvement is achieved without requiring additional training or labeling.
Even though these works have achieved commendable results, they are solely focused on classification tasks and the interpretability of models. 
Drawing inspiration from these models, we propose a training-free approach for text-supervised semantic segmentation. By employing an LLM, we can automatically generate detailed subclass descriptions, enhancing the distinctiveness of the textual features.

\noindent\textbf{CLIP-based semantic segmentation.} CLIP-based semantic segmentation~\cite{extractclip, li2022languagedriven, ghiasi2022scaling, xu2022simple, liu2022open, kirillov2023segment} utilizes text descriptions to direct image segmentation tasks, thereby mitigating the necessity for extensive manual labeling. Language-driven segmentation employs a text encoder to calculate embeddings of descriptive input labels in conjunction with a transformer-based image encoder. This image encoder is tasked with computing dense, per-pixel embeddings of the input image. ViL-Seg~\cite{liu2022open} introduced an open-world semantic segmentation pipeline. This method represents the first attempt to learn the segmentation of a wide range of semantic objects in open-world categories without needing dense annotations. It achieves this by solely utilizing image-caption data that is readily available on the Internet. Segment anything~\cite{kirillov2023segment} represents an effort to elevate image segmentation to align with the era of foundation models. This initiative has introduced a new task (promotable segmentation), developed a novel model (SAM), and compiled a unique dataset (SA-1B), all contributing to making this advancement feasible.
GroupViT~\cite{GroupViT} uses textual descriptions to assist image segmentation, yet it faces challenges related to backbone flexibility and maintaining semantic consistency, mainly due to the inherent ambiguity of textual guidance.
Simseg~\cite{SimSeg} is a method that emphasizes the significance of non-contextual and contextual information for semantic segmentation by introducing a strategy called Locality-Driven Alignment (LoDA) to prevent over-reliance on optimization with only contextual information.
Although these studies have achieved remarkable results, they have not explored using subclass descriptors to attain more detailed segmentation outcomes. Our paper utilizes detailed subclasses for each category to accomplish more thorough segmentation results without additional training.
\section{Preliminaries}
Before detailing our training-free semantic segmentation methodology, we briefly describe the CLIP-based semantic segmentation and introduce the necessary notation.

\noindent\textbf{Text-supervised semantic segmentation.} 
The primary goal of CLIP~\cite{CLIP} is to employ a contrastive pre-training approach using a substantial dataset of paired images and texts to train two encoders, namely, $f_I$ for processing image data $I$ and $g_T$ for processing text data $T$. This approach encourages these encoders to establish alignment between corresponding image-text pairs within a joint semantic space. 
Given an image $ X \in I $, the [$f_I(\mathbf X )] \in \mathbb{R}^ {{m}_i \times d} $  represent image feature, where ${m}_i$ denotes the number of visual tokens, and $d$ represents the feature dimension. Similarly, the encoded feature is $[g_T(\mathbf W)] \in \mathbb{R}^ {{m}_t \times d} $ for a text sequence $ W \in T $, with ${m}_t$ indicating the number of word tokens. 
In the context of segmentation, $Y$ denotes a sentence generated based on the object class name, serving as a prompt for the segmentation task.

For the k-th visual token, denoted as ${X}_k \in X$, and its corresponding encoded feature ${ \left [ f_I(\mathbf {X}) \right ] }_k$, the similarity scores concerning the entire image and text features are:
\begin{equation}
{S}^{t}_k \in \mathbb{R}^1 = \frac{1}{m_{t}}  \sum_{j=1}^{m_{t}}{ \left [ g_T(\mathbf {W}) \right ] }^{t}_j \times { \left [ f_I(\mathbf {X}) \right ] }_k ,
\end{equation}
Where $m_t$ is the length of word tokens. In the segmentation process, the sentence $\mathbf{W}$ is initiated by mentioning a specific object class that needs to be segmented.
The patch-wise similarity maps can be denoted as follows:
    \begin{equation}
M^{\mathrm{sim}} \triangleq \left \{  {{S}^{t}_k}  \right \} _{k=1}^{m_{i}}.
\end{equation}
Utilizing the similarity map $M^{\mathrm{sim}}$ as a basis, we can subsequently generate a categorical segmentation mask through a series of post-processing operations, including procedures such as reshaping and thresholding~\cite{SimSeg}.

\section{Methods}
\label{sec:methods}

This section details our approach to training-free semantic segmentation under LLM supervision. Initially, we describe employing an LLM (GPT-3) to generate detailed subclass descriptions in Section~\ref{sec:subclass generation}. Following this, we discuss how these subclasses are applied as target labels in an advanced text-supervised semantic segmentation model in Section~\ref{sec:subclass segmentation}. Additionally, in Section~\ref{sec:ensemble}, we suggest ensembling subclass descriptors, a technique that merges segmentation maps from various subclass descriptors with the original superclass descriptor.

\subsection{Generation of the subclass with an LLM}
\label{sec:subclass generation}
We employ the GPT-3 model~\cite{GPT3} to automatically generate detailed subclasses, which provide more informative and distinguishable features for each class. This approach also lessens the dependence on human experts for subclass generation, aligning with our objective of fully automating this process, as inspired by the concepts in ~\cite{CHILS}.
GPT-3 demonstrates substantial knowledge in subclasses, effectively representing each class when given appropriate prompts. Specifically, for a set size $n$ of subclasses and a given superclass name \texttt{class-name}, we prompt GPT-3 with $\mathcal{P}_1$:
\begin{tcolorbox}[title= $\mathcal{P}_1$:  General subclass generation guide, colframe=lightblue]
\begin{flushleft}
\texttt{Q: List \textit{n} subclasses of the following: \textbf{\textit{class-name}} \\}
\texttt{A: Here are \textit{n} commonly seen subclasses of \textbf{\textit{class-name}}:}
\end{flushleft}
\end{tcolorbox}
\noindent Here \textit{n} is the number of the generated subclass name, we set $\textit{n}=10$ in this paper. We include an experiment that examines the impact of varying the number of superclasses in the supplemental material.

To optimize performance with the above prompt, we supply GPT-3 with two examples of desired outputs for few-shot adaptation. Notably, these examples can be standardized and applied consistently across all datasets, eliminating the need for additional user input when generating a subclass set for a new dataset. We show the optimized prompt  $\mathcal{P}_2$ as follows:
\begin{tcolorbox}[title= $\mathcal{P}_2$: Subclass generation guide with few examples, colframe=lightblue]
\begin{flushleft}
\texttt{\small Q1:List 3 subclasses of the \textit{\textbf{person}}:\\}
\texttt{\small A1:female, male, child\\}
\texttt{\small Q2:List 3 subclasses of the \textit{\textbf{boat}}:\\}
\texttt{\small A2:fishing boat, cruise ship, ship\\}
\texttt{\small Q:List \textit{n} subclasses of the following: \textbf{\textit{class-name}}  \\}
\texttt{\small A:Here are \textit{n} commonly seen subclasses of \textbf{\textit{class-name}}: }
\end{flushleft}
\end{tcolorbox}
After acquiring the subclasses generated by GPT-3, we aim to use these subclasses to supervise each image for more accurate predictions. The next section will discuss applying these subclasses in an advanced text-supervised semantic segmentation approach.

\begin{figure}[t]
    \centering
    \includegraphics[width=1.\textwidth]{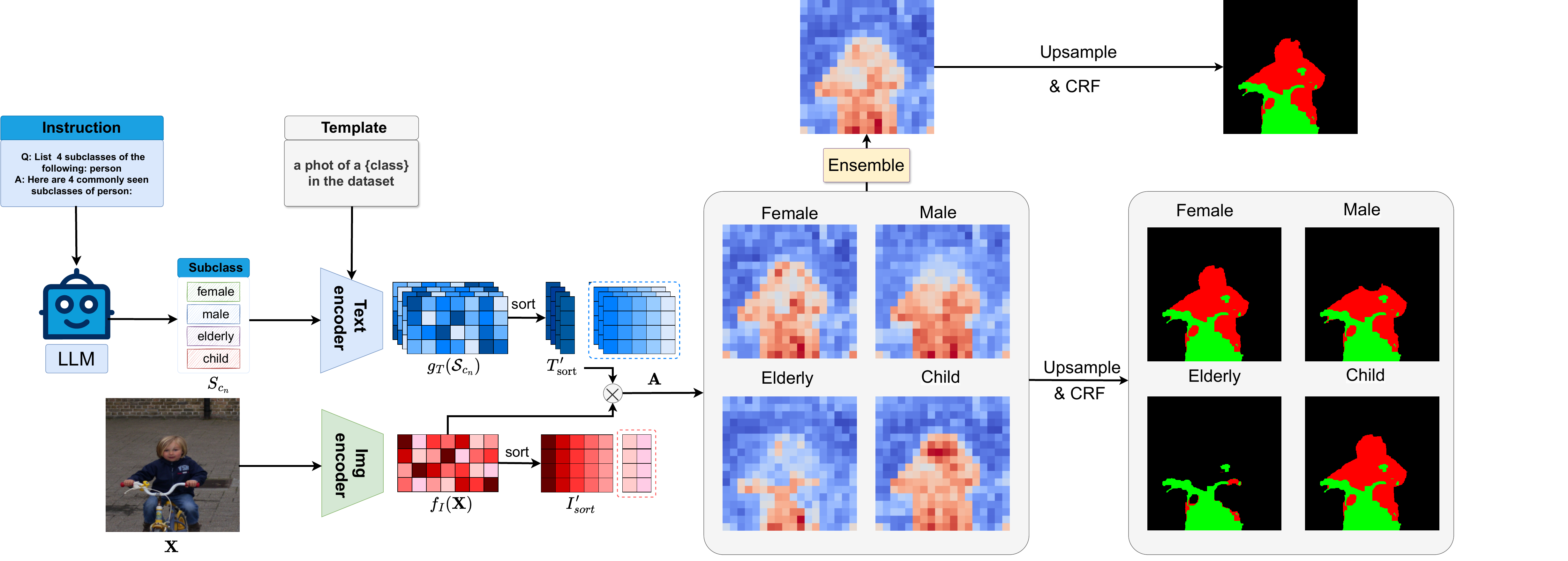}
    \caption{\textbf{Training-free semantic segmentation via LLM supervision.} Initially, the LLM generates four subcategories within the `person' category. These generated prompts and a template are then processed through a text encoder to derive textual features, while image features are extracted using an image encoder. The next step involves calculating the similarity $\mathbf{A}$ between the textual features, which are organized and aligned with a locality-driven approach, and the image features. This similarity is utilized to create rough masks for each subclass. For greater accuracy, these masks are refined using upsampling and a conditional random field (CRF). Finally, an ensemble technique is employed to merge the rough masks of the four subclasses, resulting in the final prediction.}
    \label{fig:framework}
\end{figure}

\subsection{Training-free semantic segmentation}
\label{sec:subclass segmentation}
We leverage SimSeg~\cite{SimSeg} as our baseline, the state-of-the-art data-free text-supervised semantic segmentation. Each superclass label, denoted as $c$, represents a specific concept in natural language, \eg, \textit{person}. We then input each superclass label $c$ into the GPT-3 to generate a list of subclass set $\mathcal{S}_{c_n}$, \eg, $\{\textit{female},  \textit{male}, \textit{elderly}, \textit{child}\} $, where $c_n$ is the $n-th$ subclass name of superclass $c$.
We enter all the subclasses' names generated into the text encoder using specific predefined templates, such as \textit{a photo of a \{subclass\}}. The features of the generated subclass are represented as $[g_T(\mathcal{S}_{c_n})] \in \mathbb{R}^ {n \times {m}_t \times d}$. 

Next, we input the test image $X$ into the image encoder to extract image features, denoted as $f_I(\mathbf X ) \in \mathbb{R}^ {{m}_i \times d}$. In line with SimSeg~\cite{SimSeg}, we adopt the locality-driven alignment (LoDA) technique using the feature selection method of maximum response selection. This method helps avoid the overly dense alignment of pixels and entities during the optimization of CLIP. The textual features are then arranged in descending order along the dimension $d$:
\begin{equation}
{T}_{\mathrm{sort}} \in \mathbb{R}^ {n \times {m}_t \times d}  = \mathrm{sort}_d([g_T(\mathcal{S}_{c_n})]).
\end{equation}
By doing so,  LoDA selects the features that exhibit the highest values in each channel as ${T}'_{\mathrm{sort}} \in \mathbb{R}^ {n \times d}$, anticipated to encompass vital local textual concepts. Subsequently, we determine the resemblance between image features $f_I(\mathbf X ) $, and sorted text features ${T}'_{\mathrm{sort}}$, to derive the attention weight $\mathbf{A}$. This weight will play a crucial role in the ensemble phase. The calculation of the attention weight $\mathbf{A}$ is executed as follows: 
\begin{equation}
\mathbf{A} \in \mathbb{R}^ {n \times {m}_i} = [f_I (\mathbf X )] \times {T}'_{\mathrm{sort}}.
\end{equation}
Subsequently, we generate an $n$ initial coarse mask by applying attention weight $\mathbf{A}$ to each original image pixel. 
%
% To obtain the more precise masks, we adopt a series of post-processing procedures, which also not need the additional training. 
% First, we use upsampling techniques to carefully recover complex details in the image. This ensures the restoration of lost information, leading to an overall enhancement in the visual accuracy of the upsampled features. 
% Then, we apply the conditional random field (CRF)~\cite{CRF} as a critical step in optimizing the results of upsampling.
% By integrating the CRF into the process, we consolidated the enhanced features obtained through upsampling, facilitating the capture and utilization of extended pixel relationships. 
% This integration helps a comprehensive understanding of the intricate structures within the image. 
% we obtained finely segmented results for $n$ distinct subclasses within the tested superclass. 
% Next, we will explore the fusion of textual information across different subclasses. This integration aims to enhance the overall semantic analysis of the image, resulting in more precise segmentation masks.

We implement several post-processing steps to achieve more accurate mask results that do not require additional training. 
Initially, we employ upsampling techniques to meticulously restore complex details in the image. This helps recover information that might have been lost, thereby improving the visual accuracy of the upsampled features.
Subsequently, we utilize the conditional random field (CRF)~\cite{CRF} method as a crucial step to refine the outcomes of the upsampling process. The inclusion of CRF aids in solidifying the enhanced features from upsampling, enabling the effective capture and use of expanded pixel relationships. This step is pivotal in thoroughly understanding the complex structures present in the image.
Finally, we achieve finely segmented results for \(n\) distinct subclasses within the tested superclass. Following this, we delve into the fusion of textual data across various subclasses. The purpose of this fusion is to augment the overall semantic interpretation of the image, leading to segmentation masks of higher precision. The overall framework is shown in the Figure~\ref{fig:framework}. The middle four images are the coarse masks before upsampling and CRF.  
%
% With a subclass set defined for each class, we advance to the next step in training-free semantic segmentation. In this approach, we employ the \emph{union} of all subclass sets to create a comprehensive set of classes.
%
% Our model employs a pre-trained framework~\cite{SimSeg} characterized by sparse alignment of critical image pixels and essential entities. When dealing with semantically segmented images in the dataset, we convert prompts into sentences based on the number of subclass. For example, a prompt might be "a photo of a \{subclass\}." We compute the image and text features using the respective encoders of our pre-trained models.

\begin{figure}[t]
    \centering
    \includegraphics[width=0.85\textwidth]{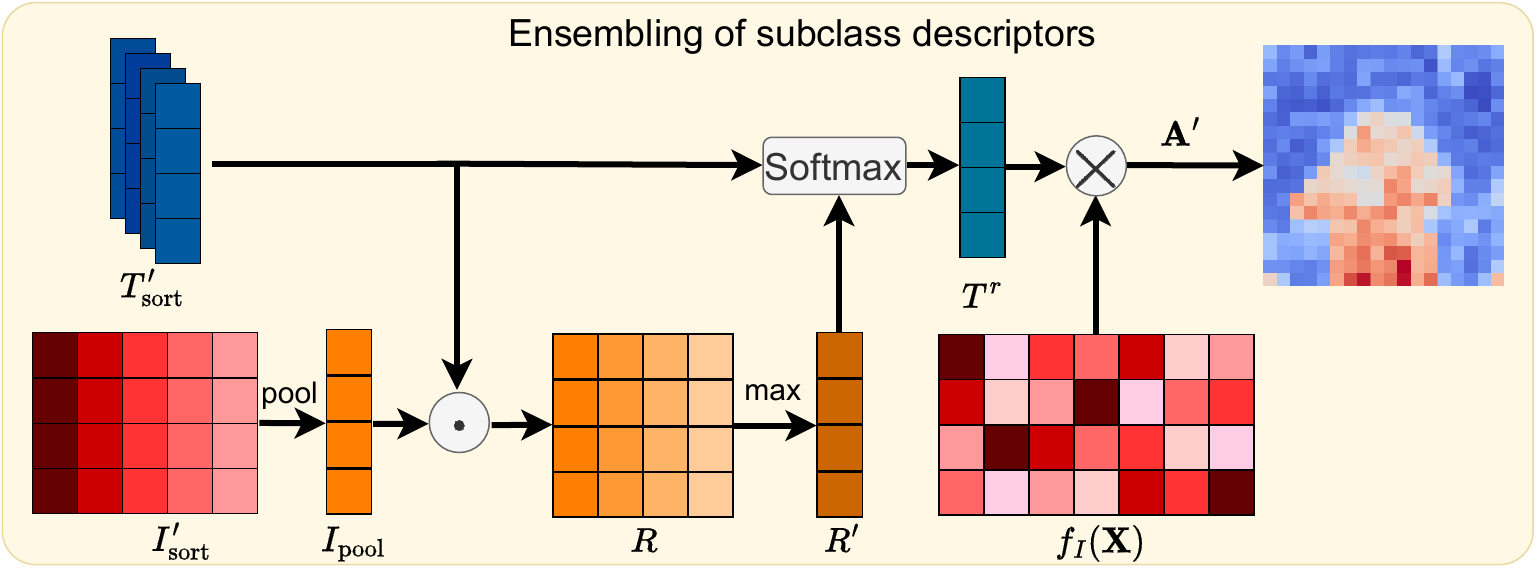}
    \caption{\textbf{Ensembling of subclasses descriptors}. We integrate descriptors from different subclasses to achieve more accurate segmentation results by utilizing the attention weights between chosen textual and visual features.}
    \label{fig:ensemble}
\end{figure}
\subsection{ Ensembling of subclasses descriptors } 
\label{sec:ensemble}
We propose subclass descriptor ensembling to merge segmentation results from various subclasses, thereby achieving more precise segmentation results. 
We initially perform feature selection on image features $[f_I(\mathbf X )] \in \mathbb{R}^ {{m}_i \times d}$:
\begin{equation}
{I} _{\mathrm{sort}}  \in \mathbb{R}^ {{m}_i \times d}  = \mathrm{sort}_d([f_I(\mathbf X )]).
\end{equation}
Differing from the selection of textual features ${T}'_{\mathrm{sort}}$, we choose locally responsive features ${I}'_{\mathrm{sort}} \in \mathbb{R}^ {5 \times d}$ from the top 5 dimensions with maximum responses, and discard the left features. This enables us to encompass important visual concepts and critical entities within the image.
We apply an average pooling operation to the image features ${I}'_{\mathrm{sort}}$. Following this, we perform element-wise multiplication (Hadamard product) on the pooled image features ${I}_{\mathrm{pool}} \in \mathbb{R}^ {d} $ and the textual features ${T}'_{\mathrm{sort}}$, yielding relationship weights $R  \in \mathbb{R}^ {n \times d }$ between them:
\begin{equation}
R  \in \mathbb{R}^ {n \times d } = {I}_{\mathrm{pool}}  \cdot {T}'_{\mathrm{sort}}.
\end{equation}
Next, we calculate the row-wise maximum in the relationship features $R$ to obtain $ R'\in \mathbb{R}^ {n}$, which fuses important information from descriptors of different subclasses. Then, we apply the softmax operation to $R'$ and the textual features ${T}'_{\mathrm{sort}}$. Consequently, we obtain the final fused textual features ${T}^r \in \mathbb{R}^ {d}$, which are generated based on the alignment between the image features and the textual features from different subclasses.
Finally, we calculate the similarity between the textual features ${T}^r$, which contain rich subclass semantic information, and the image features $[f_I(\mathbf X )]$, to obtain the final attention weights $\mathbf{A}'$: 
\begin{equation}
\mathbf{A}' \in \mathbb{R}^ {{m}_i} = [f_I(\mathbf X )]  \times {T}^r .
\end{equation}
Using the $\mathbf{A}'$, we perform upsampling and CRF post-processing operations, ultimately generating a more precise mask. 
Our ensemble process is simple and efficient, requiring no parameter updates, thereby achieving a training-free semantic segmentation approach. We show the framework of ensembling of subclass descriptors in Figure~\ref{fig:ensemble}.

\section{Experiments}
\label{sec:experiments}

\subsection{Datasets, baselines and implementation details}

\paragraph{Datasets.} to perform evaluations of text-supervised semantic segmentation, we utilize the validation subsets of three additional datasets: (1) PASCAL VOC 2012~\cite{pascalvoc}, which focuses on object recognition and semantic segmentation, includes 20 object categories and is widely recognized as a standard benchmark; (2) PASCAL Context~\cite{pascalvoccontext}, an expansion of PASCAL VOC 2012, enhances its category list to 59 with more detailed labels for more nuanced differentiation;  (3) COCO-Stuff~\cite{cocostuff}  building upon Microsoft COCO~\cite{lin2014microsoft}, places an emphasis on detailed semantic segmentation in 80 categories, offering a more comprehensive understanding of scene semantics.

\paragraph{Baselines.}  In our comparative evaluation, we utilize several well-established baseline models: (1) A supervised model, DeiT~\cite{DeiT}, which utilizes ground truth during its training phase; (2) Self-supervised models like DINO~\cite{DINO} and MoCo~\cite{MoCo_v3}, which are pre-trained on various datasets;  (3) Text-supervised models, Group ViT~\cite{GroupViT} and SimSeg~\cite{SimSeg}, pre-trained with image-text pairs and then applied to semantic segmentation in a zero-shot manner.

\paragraph{Implementation Details.} In alignment with the procedures of SimSeg, we do not employ any augmentation techniques during the zero-shot semantic segmentation evaluation other than resizing images to a resolution of  $288 \times 288$. Our method's efficacy is assessed using the COCO-Stuff~\cite{cocostuff} and PASCAL dataset. Consistent with established protocols, we utilize 80 classes of foreground objects. To obtain the semantic segmentation mask for each image, instance masks belonging to the same category are merged. We apply the two most prevalent types of pairwise potentials: Gaussian and bilateral, each with their standard parameters~\cite{CRF}. The Gaussian potential examines spatial similarities among pixels, focusing on their proximity, while the bilateral potential accounts for both color and spatial similarities among pixels. The inference process is followed, and the label with the highest energy is selected as the final segmentation result. The output is then resized to the original image dimensions. The inference process is carried out over three iterations. Our code will be made available to the public.
 
\begin{table}[t]
  
  \centering
  \caption{
    \textbf{Benefit of LLM-supervison} for text-supervised semantic segmentation.The mIoU results on PASCAL VOC~\cite{pascalvoc}, PASCAL Context~\cite{pascalvoccontext} and COCO-Stuff~\cite{cocostuff} datasets are reported. Our LLM supervision consistently achieves better performance than all baselines on all datasets. Furthermore, the performance is notably enhanced when employing $\mathcal{P}_2$ 
  for text-supervised semantic segmentation.}
    \vspace{5mm}
  \scalebox{0.65}{
  \begin{tabular}{c|ccc|c|ccc}
  & \multicolumn{3}{c|}{\textbf{Pre-training}}  & & \multicolumn{2}{c}{\textbf{Transfer}} \\ 
\textbf{Backbones }      & \textbf{Models}                         & \textbf{Dataset}  & \textbf{Supervision}           & \textbf{Zero-shot }
  & \textbf{PASCAL VOC } 
  & \textbf{PASCAL Context } 
  & \textbf{COCO-Stuff}
  \\
\toprule
  ViT-S       & DeiT~\cite{DeiT} & ImageNet & class       & \xmark    & 53.0 & 35.9  & -  \\
\midrule
  ViT-S        & DINO~\cite{DINO}  & ImageNet & self        & \xmark    & 39.1 & 20.4   & - \\
  ViT-S        & DINO~\cite{DINO}  & CC3\&12M+YFCC & self        & \xmark    & 37.6 & 22.8  & - \\
  ViT-S        & MoCo~\cite{MoCo_v3}   & ImageNet & self        & \xmark    & 34.3 & 21.3  & - \\
  ViT-S        & MoCo~\cite{MoCo_v3}   & CC3\&12M+YFCC & self        & \xmark    & 36.1 & 23.0  & -  \\
  GroupViT     & GroupViT~\cite{GroupViT}   & CC3\&12M+YFCC & text        & \cmark    & 52.3 & 22.4 & 24.3 \\
  ViT-S     & SimSeg~\cite{SimSeg}   & CC3\&12M & text        & \cmark    & 56.6 & 25.8 & 27.2 \\
\midrule
    \rowcolor{mygray}
ViT-S   &    \textbf{Ours (with $\mathcal{P}_1$)}     & CC3\&12M & text        & \cmark    & {60.6} & {27.2} & {28.5} \\
    \rowcolor{mygray}
  ViT-S   &    \textbf{Ours (with $\mathcal{P}_2$)}     & CC3\&12M & text        & \cmark    & \textbf{61.7} & \textbf{27.8} & \textbf{29.1} \\
\bottomrule
  \end{tabular}}
  \label{tab:transfer}
\end{table}

\subsection{Comparative experiments}
\begin{figure}[ht]
    \centering
   \includegraphics[width=0.85\textwidth]{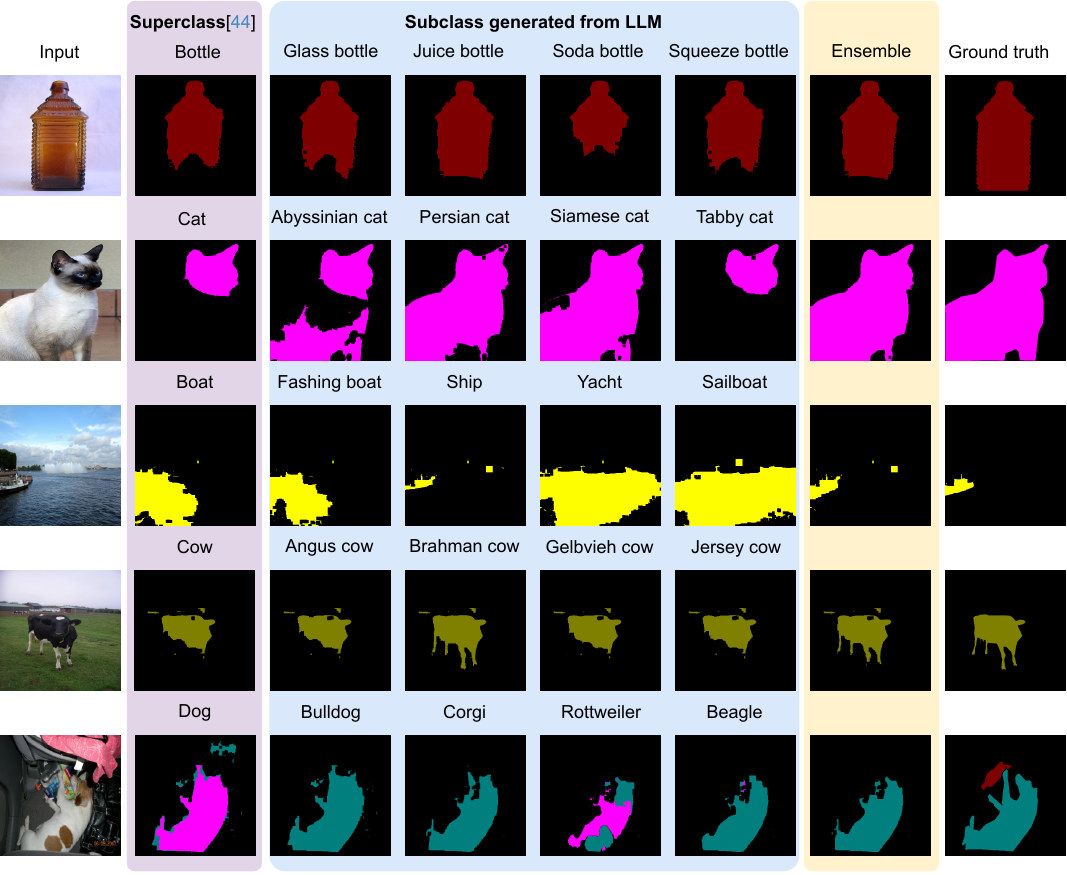}
\caption{\textbf{Segmentation Results with Our LLM-Supervision.} The use of subclass textual representations leads to more informative and precise segmentation outcomes compared to those achieved with superclass textual representations.}
\label{fig:teaser}
\vspace{-6mm}
\end{figure}

\paragraph{Benefit of LLM-supervision.}
Table~\ref{tab:transfer} compares our model and various baselines, showcasing its effectiveness. In this experiment, we apply our LLM-supervision technique to the advanced text-supervised semantic segmentation method, SimSeg~\cite{SimSeg}, which lacks image-mask pairs during training. Integrating our LLM supervision results in consistent and substantial improvement in performance across different datasets. Notably, on the PASCAL VOC dataset, our approach surpasses SimSeg by a significant 5.1\% margin. Moreover, the results using \(\mathcal{P}_2\) are consistently better than those with \(\mathcal{P}_1\). This indicates that using a few-shot sample approach can improve the generalization of the created subclasses, enhancing overall performance.
Additionally, we have included visualizations of the segmentation outcomes in Figure~\ref{fig:teaser}. The segmentation results using only superclass text-supervision tend to concentrate on the general characteristics of each class. In contrast, thanks to the generated subclasses, our LLM-supervision approach can capture more detailed information. For instance, while SimSeg, using the superclass \textit{cat}, can only segment the head of the \textit{cat}, our model successfully segments both the head and body of the \textit{cat}. 
These improvements are attributed to the LLM supervision, which generates more informative class representations, resulting in consistent enhancements over the use of superclass textual representations alone.

\begin{table}[ht]
    \centering
    \caption{\textbf{Class-wise IoUs on Pascal VOC}. The effectiveness of our LLM supervision differs among classes and is influenced by the quality of the generated subclasses.}
    \vspace{10mm}
    \scalebox{0.7}{
    \begin{tabular}{c ccccc ccccc ccccc ccccc c}
    \textbf{Models} &
    \multirow{1}{*}{\vhead{\textbf{person}}}&
    \multirow{1}{*}{\vhead{\textbf{boat}}}&
    \multirow{1}{*}{\vhead{\textbf{aeroplane}}}&
    \multirow{1}{*}{\vhead{\textbf{cow}}}&
    \multirow{1}{*}{\vhead{\textbf{monitor}}}&
    \multirow{1}{*}{\vhead{\textbf{table}}}&
    \multirow{1}{*}{\vhead{\textbf{bottle}}}&
    \multirow{1}{*}{\vhead{\textbf{horse}}}&
    \multirow{1}{*}{\vhead{\textbf{car}}}&
    \multirow{1}{*}{\vhead{\textbf{bus}}}&
    \multirow{1}{*}{\vhead{\textbf{train}}}&
    \multirow{1}{*}{\vhead{\textbf{cat}}}&
    \multirow{1}{*}{\vhead{\textbf{dog}}}&
    \multirow{1}{*}{\vhead{\textbf{bicycle}}}&
    \multirow{1}{*}{\vhead{\textbf{plant}}}&
    \multirow{1}{*}{\vhead{\textbf{bird}}}&
    \multirow{1}{*}{\vhead{\textbf{chair}}}&
    \multirow{1}{*}{\vhead{\textbf{sheep}}}&
    \multirow{1}{*}{\vhead{\textbf{motorbike}}}&
    \multirow{1}{*}{\vhead{\textbf{sofa}}}&
    \multirow{1}{*}{\textbf{Avg}}\\
    \midrule

SimSeg & 39.0 & 45.4 & 72.6 & 67.3 & 36.2 & 26.2 & 41.6 & 69.2 & 62.8 & 74.8 & 56.9 & 79.4 & 74.3 & 34.0 & 39.7 & 80.3 & 16.8 & 74.9 & 70.9 & \textbf{42.4} & 56.6\\
    \rowcolor{mygray}
\textbf{Ours} &\textbf{ 50.4} & \textbf{55.4} & \textbf{82.6} & \textbf{76.7} & \textbf{44.4} & \textbf{33.9} & \textbf{48.2} & \textbf{75.4} & \textbf{68.1} & \textbf{79.3} & \textbf{61.3} & \textbf{83.6} & \textbf{78.4} & \textbf{37.2} & \textbf{42.4} & \textbf{82.9} & \textbf{19.2} & \textbf{77.1} & \textbf{71.1} & 38.6 & \textbf{61.7}\\

$\Delta$ & \textcolor{blue}{+11.4} & \textcolor{blue}{+10.0} & \textcolor{blue}{+10.0} & \textcolor{blue}{+9.4} & \textcolor{blue}{+8.2} & \textcolor{blue}{+7.7} & \textcolor{blue}{+6.6} & \textcolor{blue}{+6.2} & \textcolor{blue}{+5.3} & \textcolor{blue}{+4.5} & \textcolor{blue}{+4.4} 
& \textcolor{blue}{+4.2} & \textcolor{blue}{+4.1} &   \textcolor{blue}{+3.2} & \textcolor{blue}{+2.7} & \textcolor{blue}{+2.6} & \textcolor{blue}{+2.4} & \textcolor{blue}{+2.2} & \textcolor{blue}{+0.2} & \textcolor{red}{-3.8} & \textcolor{blue}{+5.1} \\

    \bottomrule
    \end{tabular}}
\label{tab:specialists}
\vspace{-5mm}
\end{table}

\paragraph{Subclass quality impact. }
In our study, we used subclass text as supervision to enhance our model's performance in segmentation. The effectiveness of this approach is evident in the per superclass results on the Pascal VOC dataset, as shown in Table~\ref{tab:specialists}. With LLM supervision, our model outperforms SimSeg~\cite{SimSeg} significantly in average results. This improvement varies across classes, depending on the quality of the generated subclasses.
A notable example is the \textit{person} superclass, where our model achieves a 50.4\% score. This is attributed to the distinct features of subclasses like \textit{male}, \textit{female}, and \textit{child}.  However, our model underperforms for classes like \textit{sofa} due to fewer distinct subclasses. We have also included detailed subclass results for each superclass in our supplemental materials.
Future efforts will focus on refining subclass quality to enhance our model's efficiency and applicability. Our findings confirm that the quality of generated subclasses is critical for effective text-supervised semantic segmentation.

\paragraph{Benefit with ensembling of subclasses descriptors.}
Our study showcases 
the superiority of our ensembling of subclass descriptors is demonstrated by contrasting it with three different methods: average, cross-attention, and
\begin{wraptable}{r}{6.6cm}
\caption{\textbf{Comparison with different ensemble methods} on different datasets. Our method of ensembling subclass descriptors consistently outperforms other ensemble methods regarding mIoU.}
\vspace{-3mm}
\centering
\scalebox{0.55}{
\begin{tabular}{lc ccc}
\midrule
\textbf{Ensemble methods} & \textbf{PASCAL VOC} & \textbf{PASCAL Context} & \textbf{COCO-Stuff}\\ 
\midrule
Average & 60.1 & 27.3 & 28.6 \\
Cross-attention & 58.1 & 26.3 & 27.8 \\
Maximum similarities & 60.4 & 26.7 & 28.1 \\
\midrule
\rowcolor{mygray}
\textbf{Ours} & \textbf{61.7} & \textbf{27.8} & \textbf{29.1}\\
\bottomrule
\end{tabular}}
\vspace{-8mm}
\label{tab:comp-slowp}
\end{wraptable}
maximum similarities. As detailed in Table~\ref{tab:comp-slowp}, in the average ensemble approach, each subclass receives
equal weight, with identical importance assigned to each subclass. 
The cross-attention ensemble method applies cross-attention to each subclass descriptor. Regarding maximum similarities, we select the highest similarities between each subclass descriptor and the chosen image representation. Our ensembling subclass descriptors consistently outperform other techniques across all three datasets. Our ensembling of subclass descriptors ensures comprehensive coverage, allowing the model to effectively capture the test image's diverse aspects effectively.

\begin{wrapfigure}{r}{0.45\linewidth}
    \vspace{-9mm}
    \centering
    \includegraphics[width=0.45\textwidth]{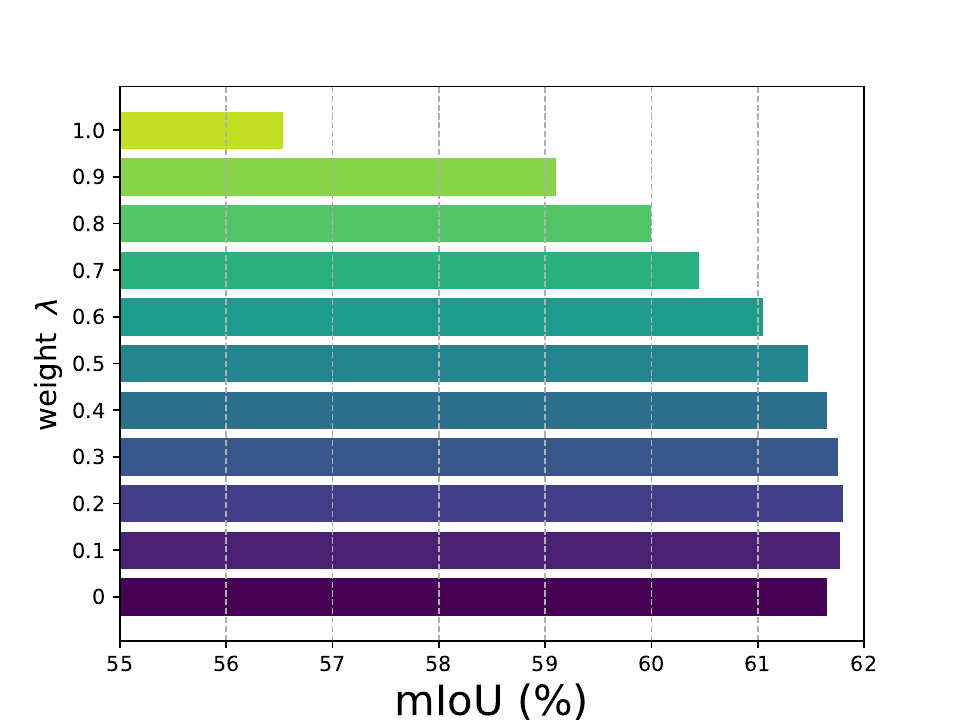}
\caption{\textbf{Impact of superclass importance}.}
    \label{fig:different_superclass_weight}
 \vspace{-9mm}
\end{wrapfigure}

\paragraph{Effect of the importance of superclass.}
In our paper, we evaluated the impact of superclass importance within our ensemble method, which incorporates the superclass textual representation for predicting final segmentation outcomes.
The analysis, depicted in Figure~\ref{fig:different_superclass_weight}, focuses on the weight given to the superclass
textual representation. The results show an increase in performance as the
weight \(\lambda\) decreases, with \(\lambda = 0.2\) yielding the optimal performance. This trend is logical, considering our generated subclasses encompass most superclass features. Significantly, the weakest performance occurs at \(\lambda = 1.0\), highlighting the advantages of including some superclass representation in our LLM-supervised model. This finding underscores the Benefit of integrating some superclasses' textual representation to enhance semantic segmentation through LLM supervision.

\paragraph{Effect of the number of generated subclasses.}
In our study, we also
\begin{wrapfigure}{r}{0.45\linewidth}
    \vspace{-8mm}
    \centering
    \includegraphics[width=0.45\textwidth]{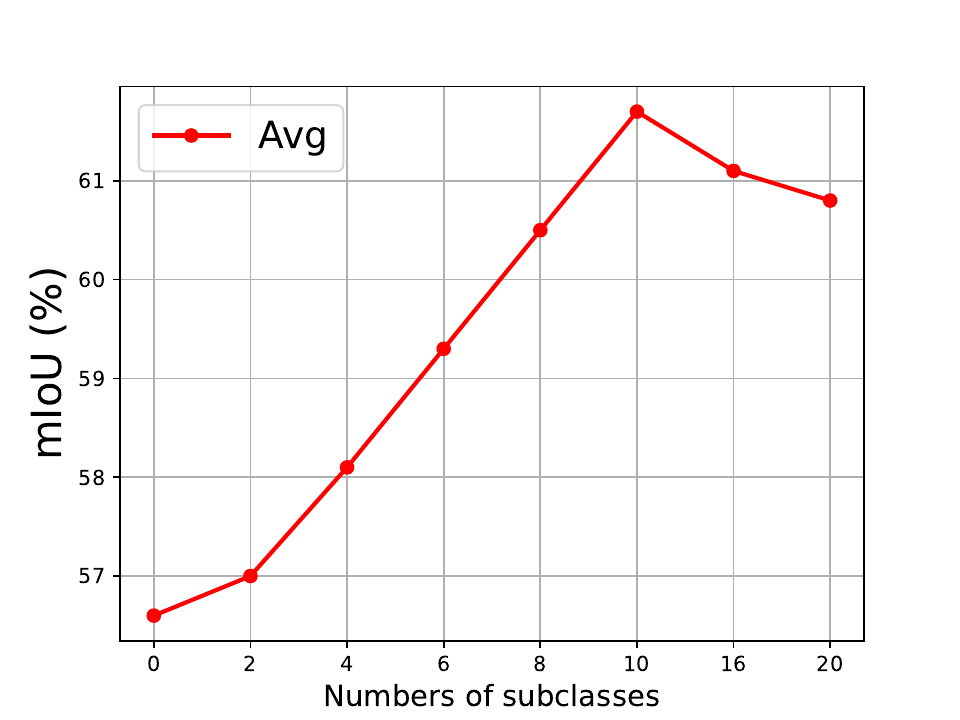}
\caption{\textbf{Effect of the number of generated subclasses.} }
    \label{fig:final_miou}
 \vspace{-9mm}
\end{wrapfigure}
analyzed
how generating different numbers of subclasses using large language models affects segmentation accuracy. 
The figure~\ref{fig:final_miou} displays an average result, showing the segmentation accuracy across all categories on the PASCAL VOC dataset for various numbers of subclasses. The segmentation accuracy for different numbers of subclasses per category in the PASCAL VOC dataset will be detailed in the supplemental materials. As the number of generated subclasses increases, segmentation accuracy steadily improves, reaching optimal precision when the number of generated subclasses is 10. This further indicates that increasing the number of generated subclasses does not lead to continuous improvement in accuracy. These results help us understand the relationship between subclass diversity and segmentation performance in various contexts. 

\begin{wrapfigure}{r}{0.5\linewidth}
\vspace{-5mm}
    \centering
    \includegraphics[width=0.5\textwidth]{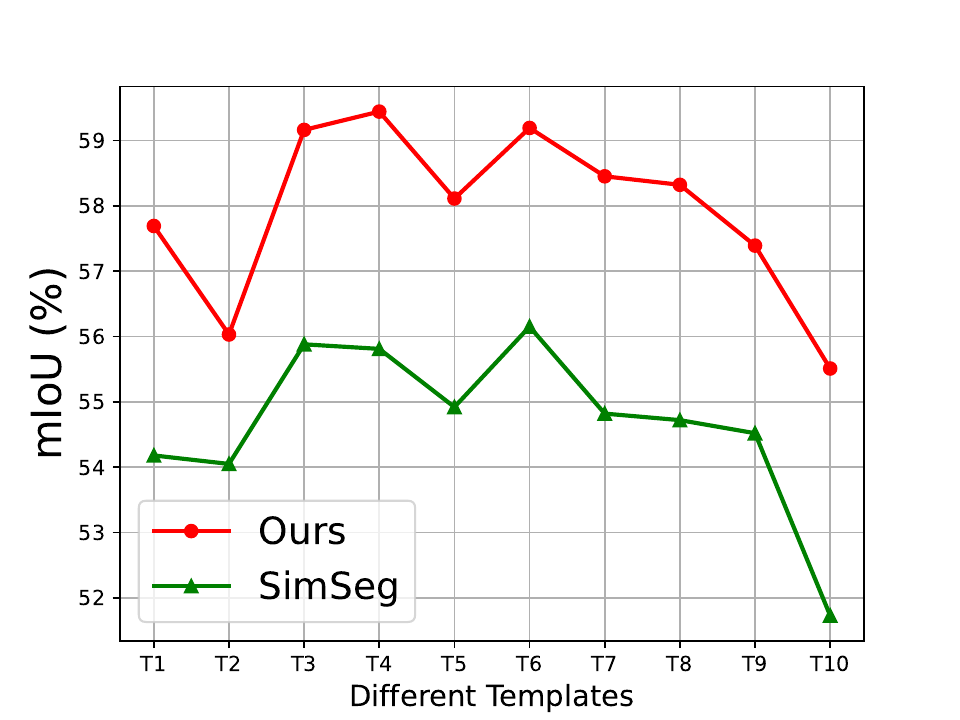}
\caption{\textbf{Influence of different templates}. Our LLM-supervised model outperforms SimSeg in text-supervised semantic segmentation, regardless of the template used.}
    \label{fig:different_templates}
 \vspace{-6mm}
 \end{wrapfigure}

\paragraph{Impact of varied template.}
The significance of prompt templates in
as highlighted in CoOP~\cite{COOP}, vision-language models led us to experiment with various templates, which serve as prompts for text-supervised semantic segmentation. These experiments are detailed in Figure~\ref{fig:different_templates}. We tested ten different templates, 
such as T1: \texttt{a drawing of a \{\}}, T2: \texttt{a photo of the cool \{\}}, among others,
with the full list provided in the supplemental materials.
Our findings reveal that performance varies with different templates. Our LLM-supervised approach consistently outperforms the SimSeg model~\cite{SimSeg}, regardless of the template used. The template T4: \texttt{a photo of a \{\}}, in particular, achieves the highest performance among the templates tested.
In this study, following the approach of SimSeg, we employed an average ensemble method to integrate representations from all templates for predicting final segmentation results. Additionally, we plan to explore prompt learning~\cite{COOP} in future work to avoid the need for manually crafted prompts.

\begin{figure}[t]
    \centering
    \includegraphics[width=0.95\textwidth]{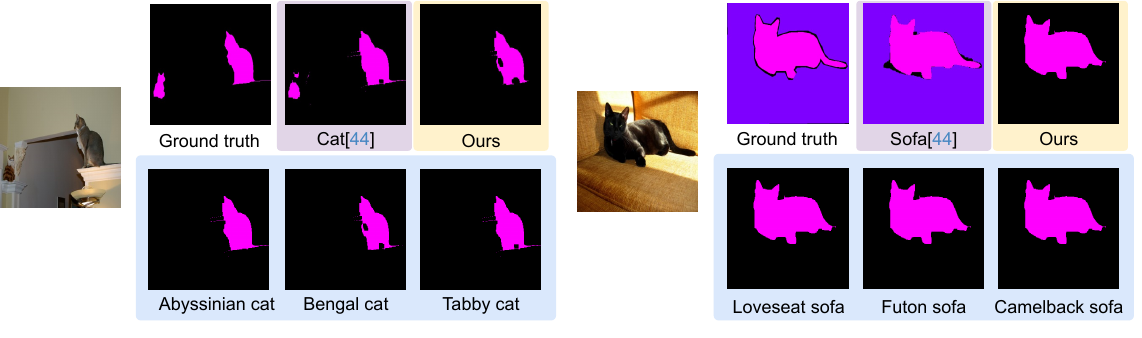}
    \caption{\textbf{Failure case with our LLM-supervision.} Our model exhibits difficulties in segmenting smaller objects and in the efficacy of the generated subclass representations.}
    \label{fig:failure}
\end{figure}

\subsection{Limitations and future work.}
The analysis of Table~\ref{tab:specialists} reveals that for the \textit{sofa} category, the performance under LLM-supervision falls short compared to SimSeg, which relies solely on superclass supervision. To delve deeper into this issue, we explore specific failure cases, as depicted in Figure~\ref{fig:failure}. Our model, guided by generated subclasses, fails to recognize the smaller cat in the first test image, featuring two \textit{cats} of varying sizes. This highlights a challenge in segmenting small objects semantically. The second test image presents a cat resting on a sofa, where our model cannot correctly segment the sofa using any of the \textit{sofa} subclasses. This suggests a limitation in the subclass quality for \textit{sofa}, likely due to these subclasses being less represented in the CLIP dataset. Addressing these two key challenges will be the focus of our future work to improve the model's generalization capabilities.

\section{Conclusions}
\label{sec:conclusions}

We introduce a novel text-supervised semantic segmentation approach leveraging large language model supervision, eliminating the need for additional training. This method begins by utilizing an LLM to generate a rich set of subclasses for enhanced class representation. 
We incorporate these subclasses as target labels into a sophisticated text-supervised semantic segmentation model, leading to a range of segmentation results that mirror the distinct characteristics of each subclass.
Further, we propose ensembling subclass descriptors to merge segmentation maps derived from different subclass descriptors, ensuring comprehensive coverage of diverse aspects within test images. Our approach's adaptability, ease of integration, and effectiveness make it a valuable addition to existing zero-shot semantic segmentation frameworks. 
We perform comprehensive ablation studies to illustrate how our method accurately captures more precise class-aware features with the generated subclasses, thus enhancing generalization.
Through comprehensive experiments on three standard benchmarks, our method demonstrates improved performance compared to traditional text-supervised semantic segmentation methods.

\section*{Acknowledgment}
This work is financially supported by the Inception Institute of Artificial Intelligence, the University of Amsterdam and the allowance 
Top consortia for Knowledge and Innovation (TKIs) from the Netherlands Ministry of Economic Affairs and Climate Policy.

\clearpage  % TODO REVIEW/FINAL: This \clearpage needs to be removed from both review and camera-ready versions.

\bibliographystyle{splncs04}
\bibliography{refs}
\clearpage

\title{Supplemental Materials: Training-Free Semantic Segmentation via LLM-Supervision}

\titlerunning{Training-Free Semantic Segmentation}

\author{Wenfang Sun\inst{1} \thanks{Equal contribution.}\and
Yingjun Du\inst{2,3}\thanks{Equal contribution. Work done during an internship at Cisco.} \and 
Gaowen Liu\inst{3}  \and \\
Ramana Kompella\inst{3}  \and
Cees G.M. Snoek\inst{2}}  
\authorrunning{Sun. et al.}
\institute{University of Science and Technology of China \and
AIM Lab, University of Amsterdam
\\
 \and
Cisco Research
\\
}

\maketitle

\section{Different template examples}

In this section, various templates are presented for generating textual representations, showcasing a range of styles and formats. The templates are presented in Table~\ref{tab:diff_templates}, labeled from T1 to T10, provide different instructions for image creation. For example, T1 suggests "a drawing of a \{\}", while T4 offers "a photo of a \{\}". Other variations include pixelated images (T3), cropped photos (T5, T8), and images with specific attributes like being bright (T7) or bad quality (T9). Each template allows for the insertion of a subject within the curly braces, indicating the flexibility and adaptability of these templates for diverse image generation requests. The table highlights the versatility of our model, catering to a wide array of preferences and requirements.  
\begin{table}[h]
    \centering
    \begin{tabular}{c|c}
    \toprule
         & Templates \\
         \midrule
        \textbf{T1} & a drawing of a \{\}\\
        \textbf{T2} & a photo of the cool \{\}\\
       \textbf{T3} & a pixelated photo of a \{\}\\
        \textbf{T4} & a photo of a \{\}\\
      \textbf{T5} & a cropped photo of the \{\}\\
       \textbf{T6} & a jpeg cropped photo of the \{\}\\
       \textbf{T7} & a bright photo of a \{\}\\
        \textbf{T8} & a cropped photo of a \{\}\\
     \textbf{T9} & a bad photo of the \{\}\\
       \textbf{T10} & a photo of many \{\}\\
       \bottomrule
    \end{tabular}
    \caption{\textbf{Different template examples.}}
    \label{tab:diff_templates}
\end{table}

\section{Results of diverse subclasses}

This section presents a comprehensive Table~\ref{tab:varied_sub}, detailing the results of diverse subclasses across various categories or 'superclasses'. The table is organized into different sections, each focusing on a specific superclass like \textit{Person}, \textit{Boat}, \textit{Aeroplane}, etc. Within each superclass, multiple subclasses are listed along with their respective mIoU values, a metric used to evaluate the accuracy of segmentation in image processing.
For instance, under the \textit{Person} superclass, subclasses like \textit{Female}, \textit{Male}, \textit{Child}, \textit{Teenagers}, etc., are listed with their mIoU scores, reflecting the performance of image segmentation techniques on these specific categories. Similar breakdowns are provided for other superclasses like \textit{Boat} (with subclasses like \textit{Sailboat}, \textit{Yacht}, \textit{Fishing boat}, etc.), \textit{Aeroplane} (including subclasses like \textit{Airliner}, \textit{Cargo aircraft}, \textit{Business jet}, etc.), and others including \textit{Cow}, \textit{Table}, \textit{Horse}, \textit{Bottle}, and more.
This detailed breakdown showcases the varying effectiveness of segmentation techniques across different subclasses within each superclass, highlighting the challenges and successes in accurately identifying and segmenting different objects and entities in images.

\begin{table*}
    \centering
    \scalebox{0.8}{
    \begin{tabular}{cc|cc|cc|cc|cc}
        \multicolumn{2}{c}{Person} & \multicolumn{2}{c}{Boat} & \multicolumn{2}{c}{Aeroplane} & \multicolumn{2}{c}{Cow} & \multicolumn{2}{c}{Table} \\
        \cmidrule(r){1-2} \cmidrule(lr){3-4} \cmidrule(lr){5-6} \cmidrule(lr){7-8} \cmidrule(l){9-10}
        Subclass & mIoU & Subclass & mIoU & Subclass & mIoU & Subclass & mIoU & Subclass & mIoU   \\
             \cmidrule(r){1-2} \cmidrule(lr){3-4} \cmidrule(lr){5-6} \cmidrule(lr){7-8} \cmidrule(l){9-10}
        Female&44.90&Sailboat& 41.94&Airliner&70.65&Holstein &67.19& Dining  &  24.99 \\
        Male&48.75&Yacht& 42.87&Cargo aircraft &71.47&Angus  &64.23  & Drafting  &29.62   \\
        Child&30.99&Fishing &40.17&Business jet &74.14 &Jersey  &  64.36& Coffee  & 24.58 \\
        Teenagers&27.10&Row &36.07& Military aircraft& 73.74& Hereford  & 64.13 & Study  & 26.09 \\
        Baby & 23.54 & Speed &40.27&  Ultralight aircraft & 78.48& Charolais  & 69.68 & Dressing  &20.65  \\
        Seniors& 18.49 & Motorboat & 35.43 &  Propeller-driven jet& 75.75 & Limousin &68.92 &Console   &  17.38 \\
        Adults & 35.30 & Canoe & 41.66 &   Fighter jet& 75.81 & Ayrshire  & 66.38 & Kitchen  & 24.82\\
        Young adults &42.59  & Inflatable  & 54.37 &  Commercial jetliner &75.75 &Simmental  & 65.59 & Picnic  & 28.81 \\
        Elderly& 28.83& Pontoon  & 37.92&   Autogyro &68.14 &Gelbvieh & 67.92& Reception  &  26.87 \\
        Actor & 48.83 & Ship & 47.21 &  Gyrodyne  & 72.52 & Brahman  & 68.38 &Work   &28.42 \\
\hline \hline
                \multicolumn{2}{c}{Horse} & \multicolumn{2}{c}{Bottle} & \multicolumn{2}{c}{Monitor} & \multicolumn{2}{c}{Car} & \multicolumn{2}{c}{Bus} \\
        \cmidrule(r){1-2} \cmidrule(lr){3-4} \cmidrule(lr){5-6} \cmidrule(lr){7-8} \cmidrule(l){9-10}
        Subclass & mIoU & Subclass & mIoU & Subclass & mIoU & Subclass & mIoU & Subclass & mIoU   \\
             \cmidrule(r){1-2} \cmidrule(lr){3-4} \cmidrule(lr){5-6} \cmidrule(lr){7-8} \cmidrule(l){9-10}
       Thoroughbred & 70.20& Water &38.42&Ultra-Wide &41.63 & Sedan  &63.41& City & 71.53\\
       Quarter  & 69.87& Soda  &39.74&4K & 40.47 & SUV &   62.02& School &73.10 \\
      Arabian &69.63&Perfume  &37.79&Television &40.43 & Hatchback & 62.18 & Double-decker &72.17 \\
         Appaloosa & 67.94 &Wine &38.48&Touchscreen &41.39  &Convertible  & 61.72 & Coach & 74.47\\
    Clydesdale & 69.60& Baby & 38.35 &Computer &34.24 & Coupe  &64.33  &Articulated  & 75.51\\
     Mustang & 69.47& Glass  &39.50&Gaming &36.72 & Minivan  &54.33  & Minibus &72.94 \\
      Paint   & 70.17 & Juice &41.12&LCD &37.17 & Electric  &59.64 & Sightseeing &73.17 \\
  Palomino &69.81 & Medicine  &46.21&Curved &34.99 &  Sports &  60.93&  Electric&74.65 \\
 Percheron & 68.64 &Spray &39.65& Professional &37.52  &Crossover SUV & 60.21 &Express  &75.61 \\
Shetland& 70.45& Squeeze  & 40.69 &LED &37.82  & Luxury  & 61.52  & Intercity & 74.22\\
        \hline \hline
                \multicolumn{2}{c}{Train} & \multicolumn{2}{c}{Cat} & \multicolumn{2}{c}{Dog} & \multicolumn{2}{c}{Bicycle} & \multicolumn{2}{c}{Plant} \\
        \cmidrule(r){1-2} \cmidrule(lr){3-4} \cmidrule(lr){5-6} \cmidrule(lr){7-8} \cmidrule(l){9-10}
        Subclass & mIoU & Subclass & mIoU & Subclass & mIoU & Subclass & mIoU & Subclass & mIoU   \\
             \cmidrule(r){1-2} \cmidrule(lr){3-4} \cmidrule(lr){5-6} \cmidrule(lr){7-8} \cmidrule(l){9-10}
        Electric  &57.61  & Tabby& 73.11& Corgi &57.94  &  Electric&  36.16& Rose &41.80 \\
        Subway& 57.09 &Siamese&77.75&Labrador retriever&75.51&Vintage&35.41&Sunflower& 38.54\\
        High-Speed& 56.76& Persian& 80.61& German shepherd& 70.36&Hybrid & 35.39&Tulip &38.27 \\
        Freight &  55.42&Maine coon &82.30 &Bulldog & 60.82 &Touring & 34.60 &Fern & 32.92\\
        Passenger & 55.99 & Sphynx & 52.54 & Poodle &64.72 & Road & 32.71 &Cactus& 40.84\\
        Light rail & 53.92 & Ragdoll & 80.69 &Beagle &67.50 & Moutain & 32.83 &Lavender &32.91 \\
        Monorail& 57.98 &Bengal& 68.13 &Rottweiler&55.93 & Cruiser & 34.17 &Flowering&41.21 \\
        Underground & 57.16 & Scottish fold & 80.77 & Boxer& 64.97 &Folding  & 31.05 &Orchid&30.60 \\
       Commuter &55.62  & Russian blue & 73.58 & Dachshund &54.14& BMX & 33.33 &Water& 33.67\\
        Intercity&54.34&Abyssinian&72.26&Golden retriever&69.96&Single speed&35.95&Houseplants&34.24 \\
        \hline \hline
                \multicolumn{2}{c}{Bird} & \multicolumn{2}{c}{Chair} & \multicolumn{2}{c}{Sheep} & \multicolumn{2}{c}{Motorbike} & \multicolumn{2}{c}{Sofa} \\
        \cmidrule(r){1-2} \cmidrule(lr){3-4} \cmidrule(lr){5-6} \cmidrule(lr){7-8} \cmidrule(l){9-10}
        Subclass & mIoU & Subclass & mIoU & Subclass & mIoU & Subclass & mIoU & Subclass & mIoU   \\
             \cmidrule(r){1-2} \cmidrule(lr){3-4} \cmidrule(lr){5-6} \cmidrule(lr){7-8} \cmidrule(l){9-10}
        Eagle & 79.48 & Lounge &17.83  & Merino &73.62  &Cruiser  & 70.20 & Reclining &38.89 \\
        Sparrow&74.50 & Armchair &17.06&Dorset&73.98&Sport bike&69.87& Sectional &37.23 \\
        Pigeon & 84.10 & Dining & 14.85 &Suffolk & 74.31 &Touring& 69.63 &Camelback  &35.17 \\
       Penguin& 80.91 & Office & 16.96 &Romney & 72.30 & Dirt bike & 67.97& Futon &33.51 \\
       Hummingbird& 52.56 & Rocking &16.32 &Jacob & 74.08&Chopper& 69.60 &Chesterfield &37.84 \\
      Owl&62.23&Folding&15.74&Border leicester&74.30 &Adventure&69.47&Contemporary &36.18 \\
       Flamingo& 65.26 & Wingback &19.20  & Cheviot & 70.25 & Bobber & 70.17 &Sleeper&38.77 \\
       Peacock &75.81&Club&18.15&Rambouillet&71.65&Cafe racer&69.81&Daybed& 39.51\\
       Pelican& 81.35 & Glider &17.29  & Cotswold & 73.49 &Scooter  &68.64&Loveseat&40.02 \\
       Finch& 78.09 & Recliner & 19.94 & Karakul & 71.45 &Trike  & 70.45 & Convertible &35.33 \\
        \bottomrule
    \end{tabular}}
    \caption{\textbf{Results of Diverse Subclasses on Each Superclass.} The supervision of diverse subclasses leads to varying results in segmentation.}
    \label{tab:varied_sub}
\end{table*}

\section{Effect of the number of generated subclasses}

\begin{figure*}[t]
    \centering
    % Row 1
   \begin{tabular}{cccc}
        \begin{minipage}{0.22\textwidth}
            \includegraphics[width=\linewidth]{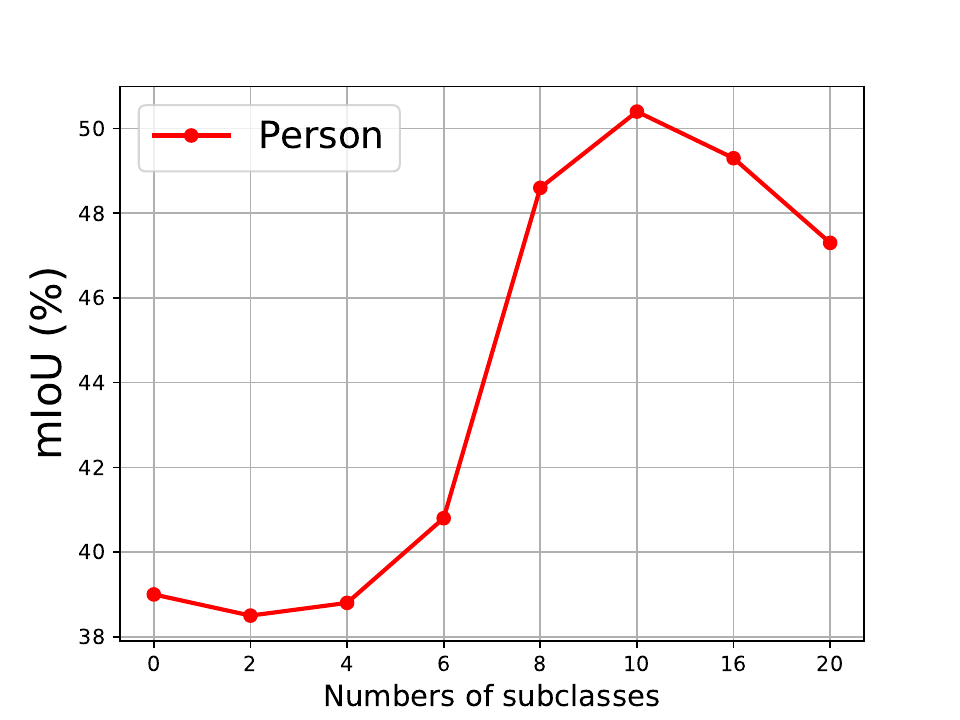}
            \caption*{Person}

        \end{minipage} &
        \begin{minipage}{0.22\textwidth}
            \includegraphics[width=\linewidth]{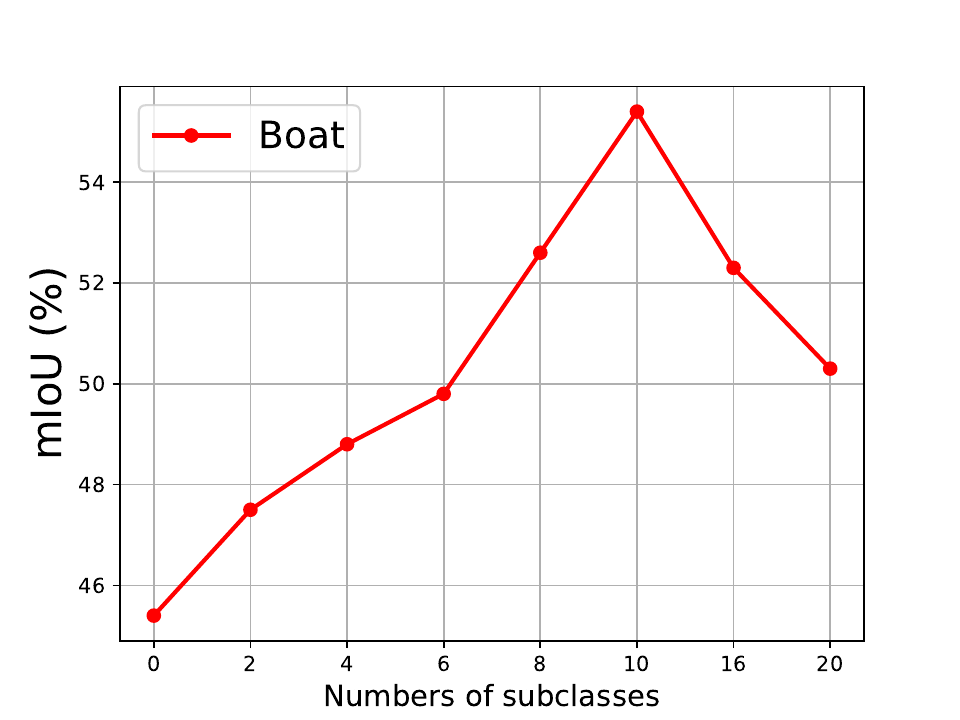}
            \caption*{Boat}

        \end{minipage} &
        \begin{minipage}{0.22\textwidth}
            \includegraphics[width=\linewidth]{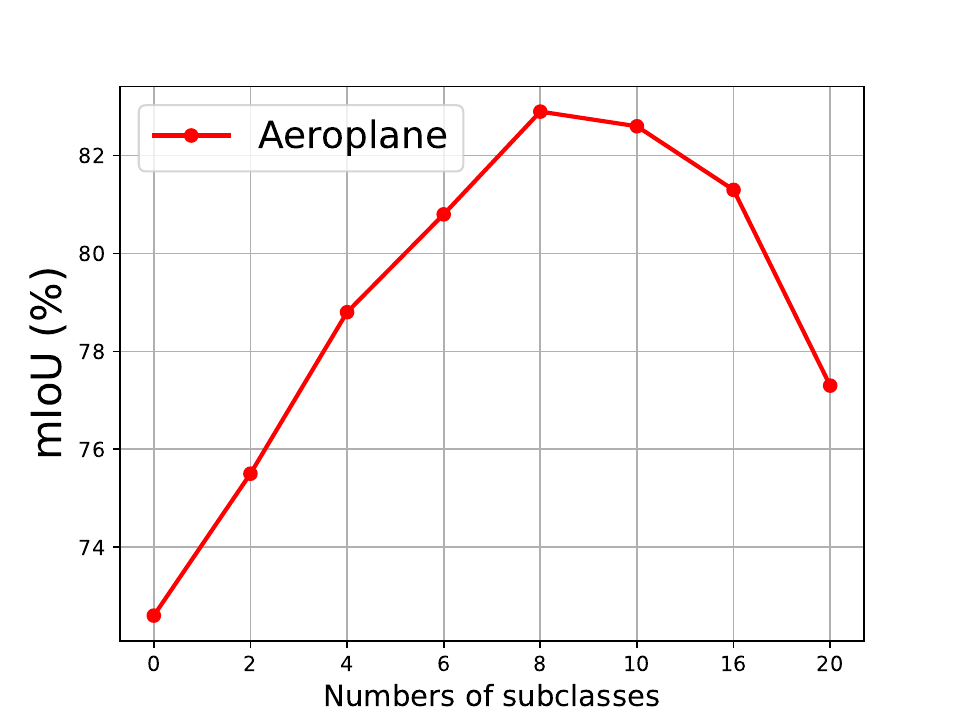}
            \caption*{Aeroplane}

        \end{minipage} &
        \begin{minipage}{0.22\textwidth}
            \includegraphics[width=\linewidth]{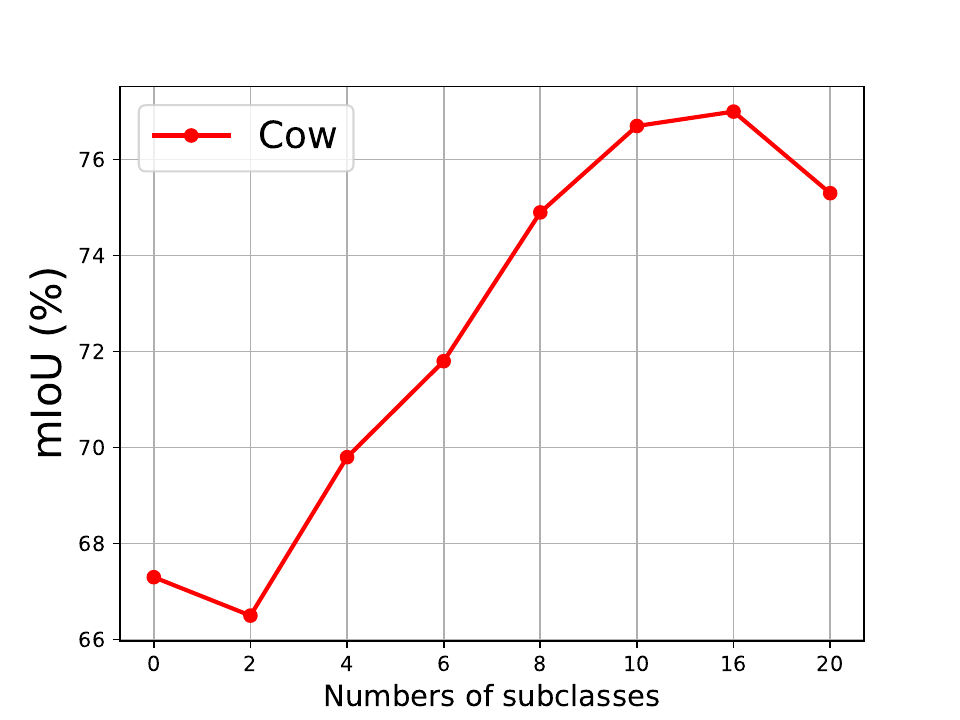}
            \caption*{Cow}

        \end{minipage} \\
    \end{tabular}

    % Row 2
   \begin{tabular}{cccc}
        \begin{minipage}{0.22\textwidth}
            \includegraphics[width=\linewidth]{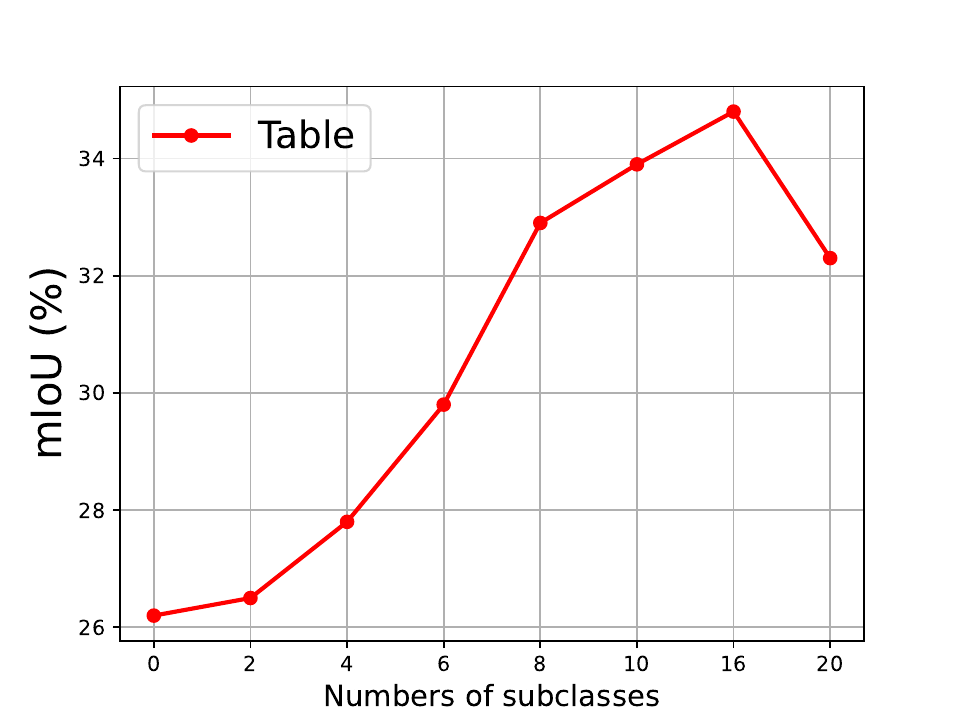}
            \caption*{Table}

        \end{minipage} 
        &
        \begin{minipage}{0.22\textwidth}
            \includegraphics[width=\linewidth]{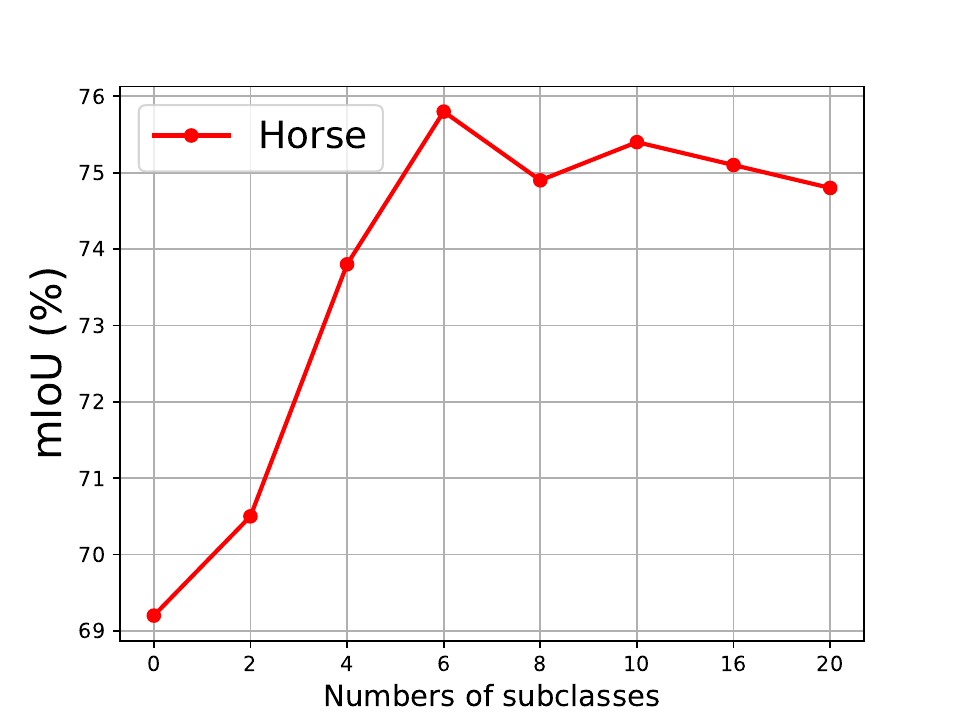}
            \caption*{Horse}

        \end{minipage}
        &
                \begin{minipage}{0.22\textwidth}
            \includegraphics[width=\linewidth]{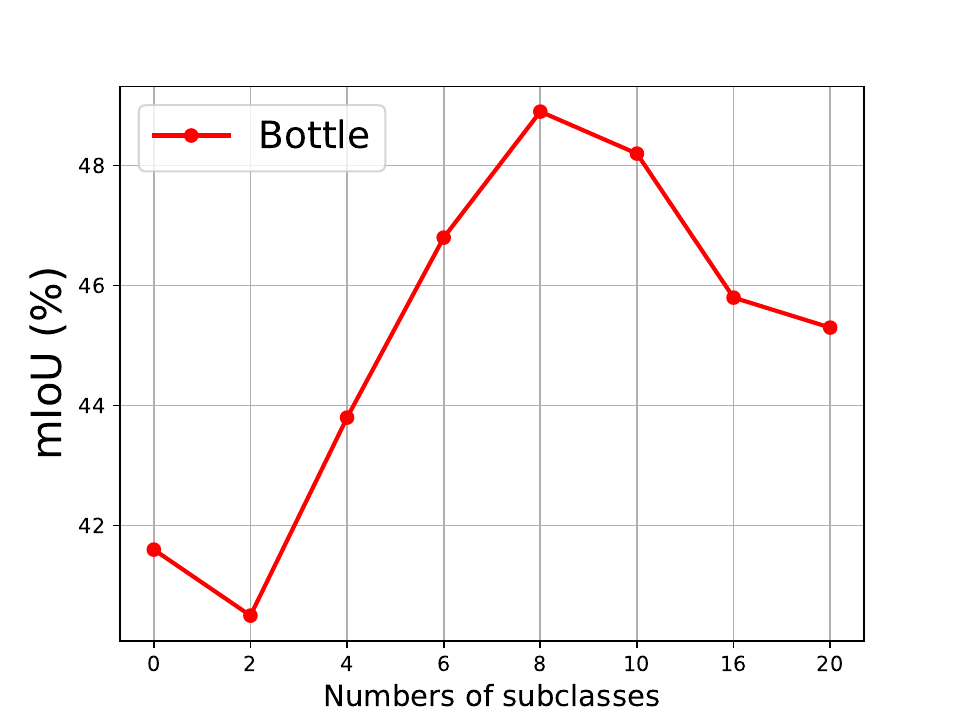}
            \caption*{Bottle}

        \end{minipage} 
        &
        \begin{minipage}{0.22\textwidth}
            \includegraphics[width=\linewidth]{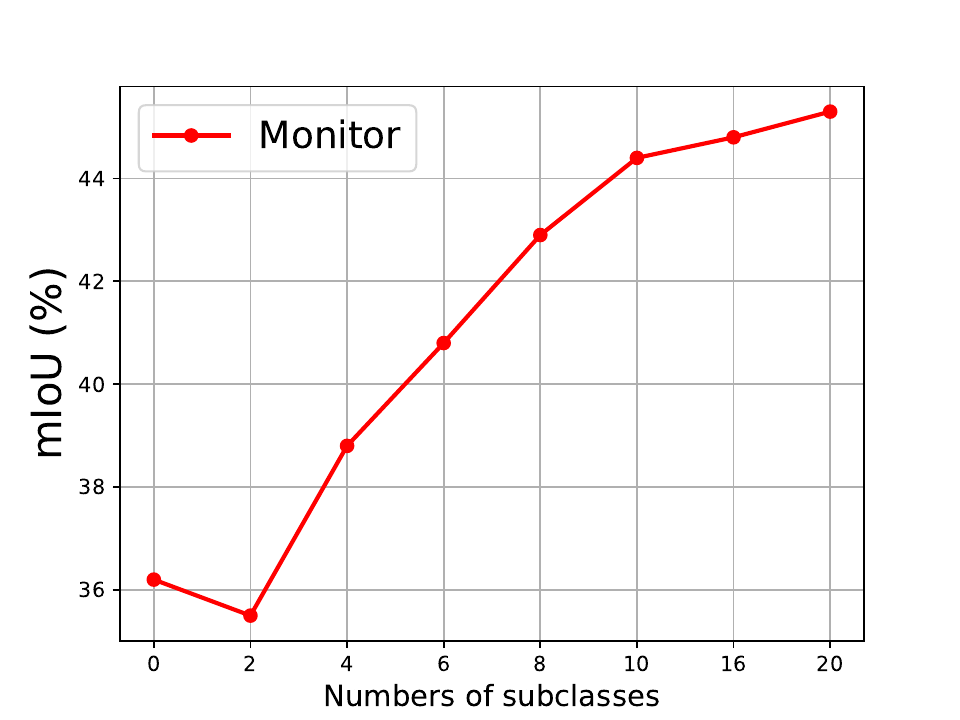}
            \caption*{Monitor}

        \end{minipage} 
 \\
    \end{tabular}

    % Row 3
   \begin{tabular}{cccc}
        \begin{minipage}{0.22\textwidth}
            \includegraphics[width=\linewidth]{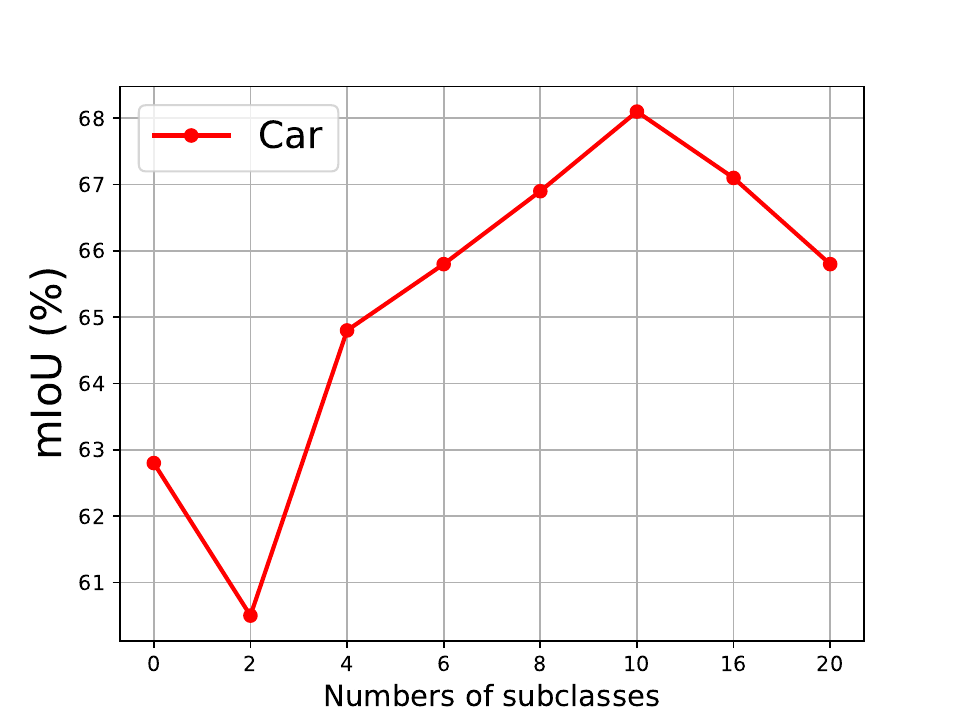}
            \caption*{Car}

        \end{minipage} &
        \begin{minipage}{0.22\textwidth}
            \includegraphics[width=\linewidth]{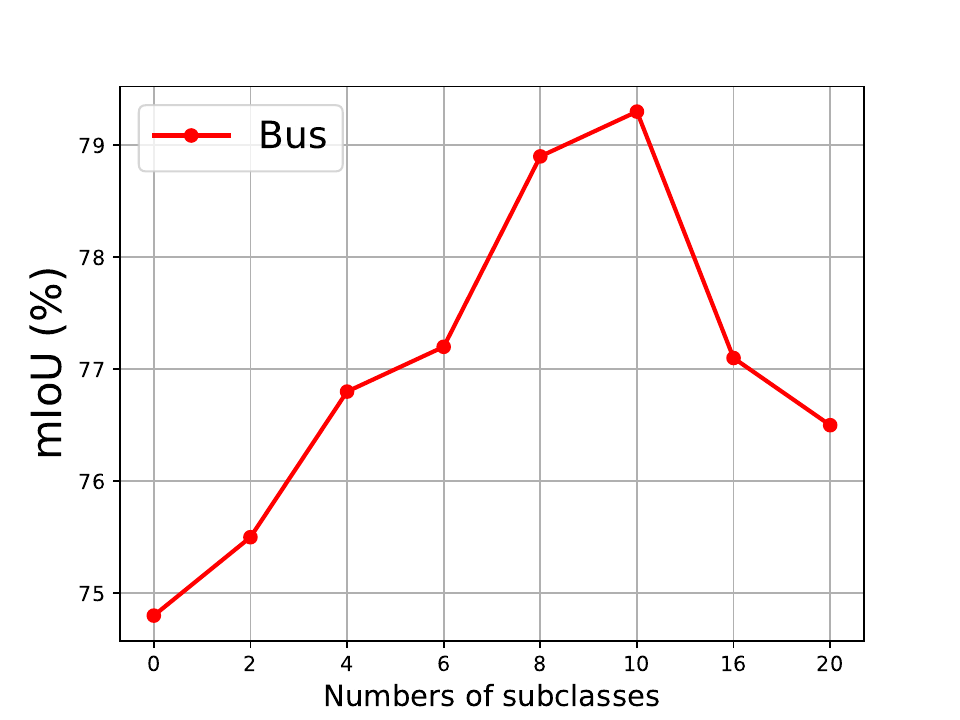}
            \caption*{Bus}

        \end{minipage} &
        \begin{minipage}{0.22\textwidth}
            \includegraphics[width=\linewidth]{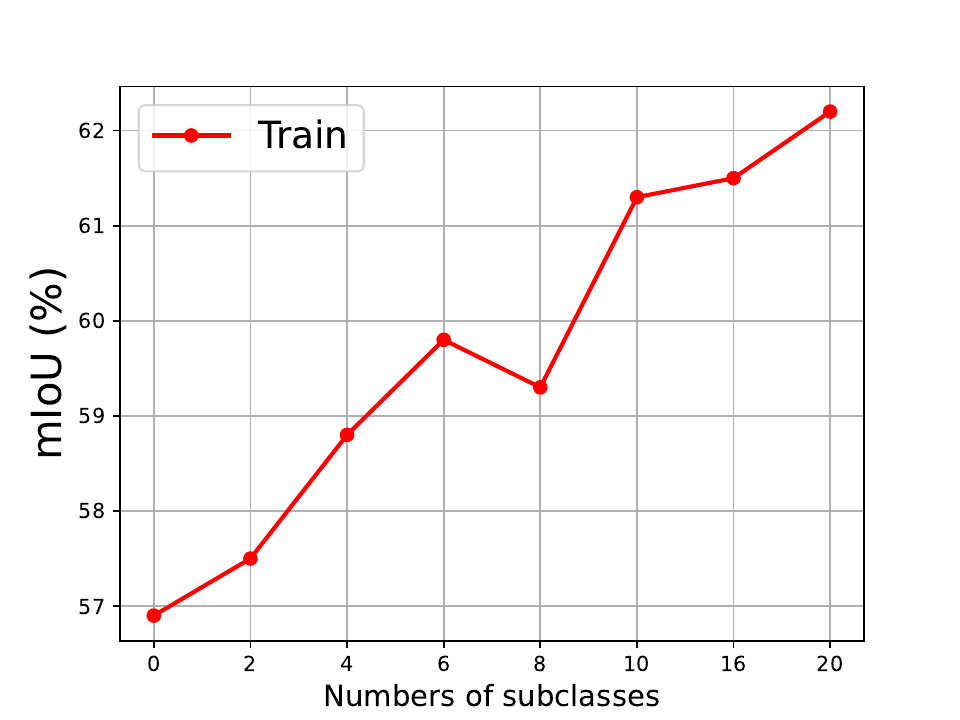}
            \caption*{Train}

        \end{minipage} &
        \begin{minipage}{0.22\textwidth}
            \includegraphics[width=\linewidth]{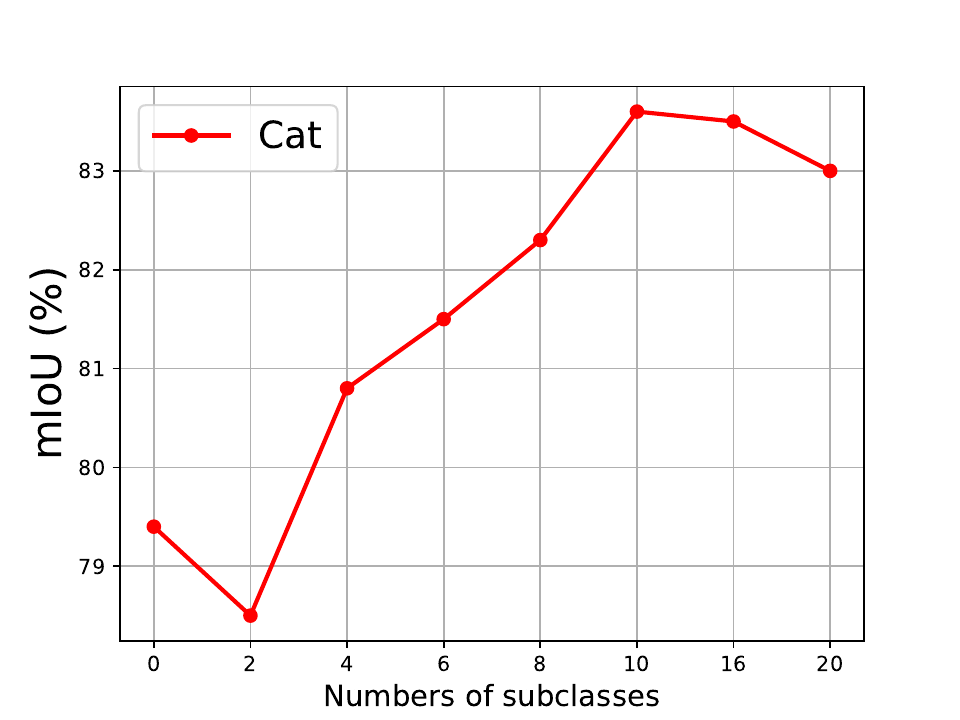}
            \caption*{Cat}

        \end{minipage} \\
    \end{tabular}

    % Row 4
   \begin{tabular}{cccc}
        \begin{minipage}{0.22\textwidth}
            \includegraphics[width=\linewidth]{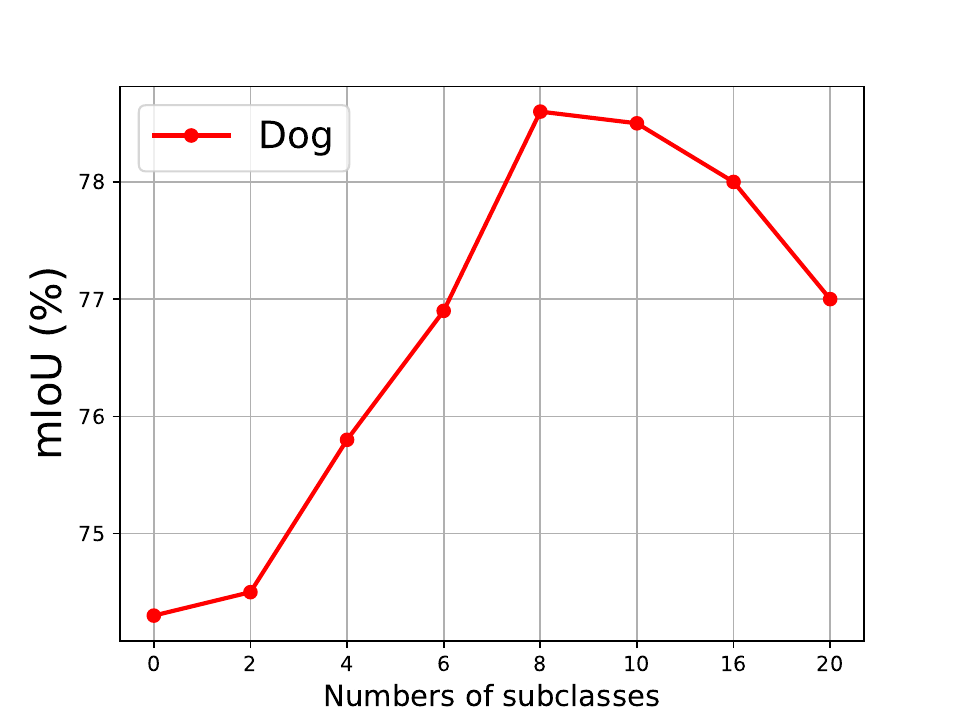}
            \caption*{Dog}

        \end{minipage} &
        \begin{minipage}{0.22\textwidth}
            \includegraphics[width=\linewidth]{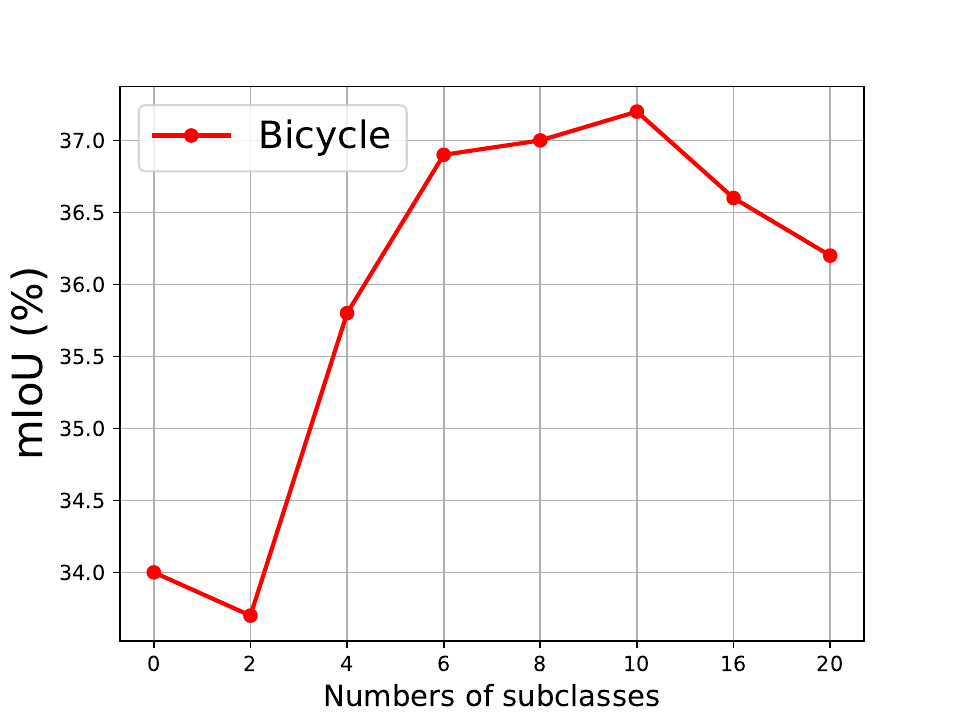}
            \caption*{Bicycle}
        \end{minipage} &
        \begin{minipage}{0.22\textwidth}
            \includegraphics[width=\linewidth]{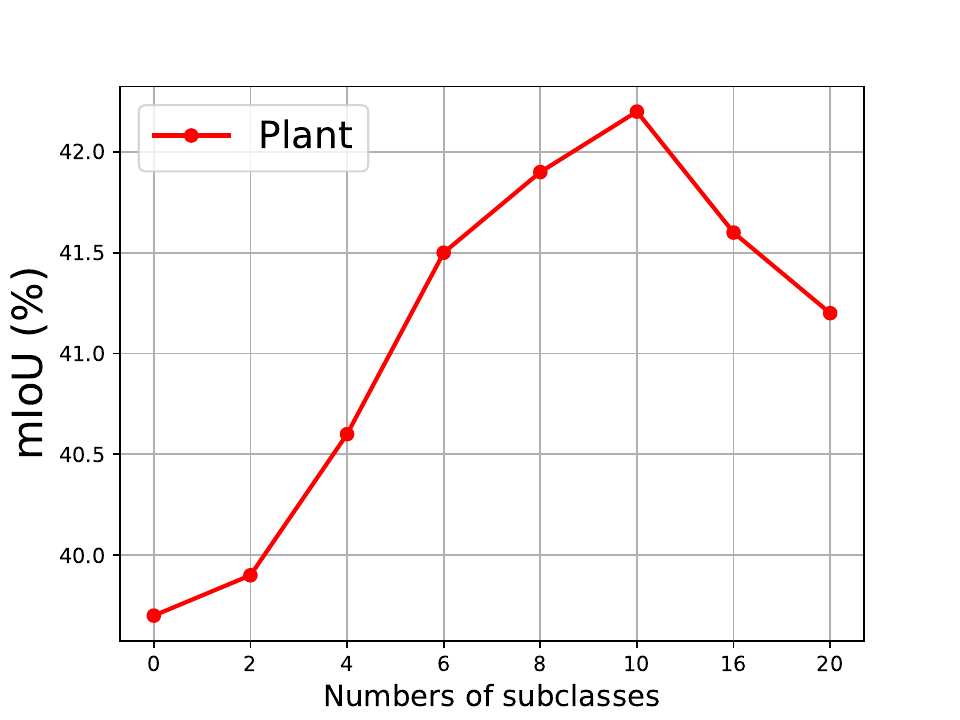}
            \caption*{Plant}
        \end{minipage} &
        \begin{minipage}{0.22\textwidth}
            \includegraphics[width=\linewidth]{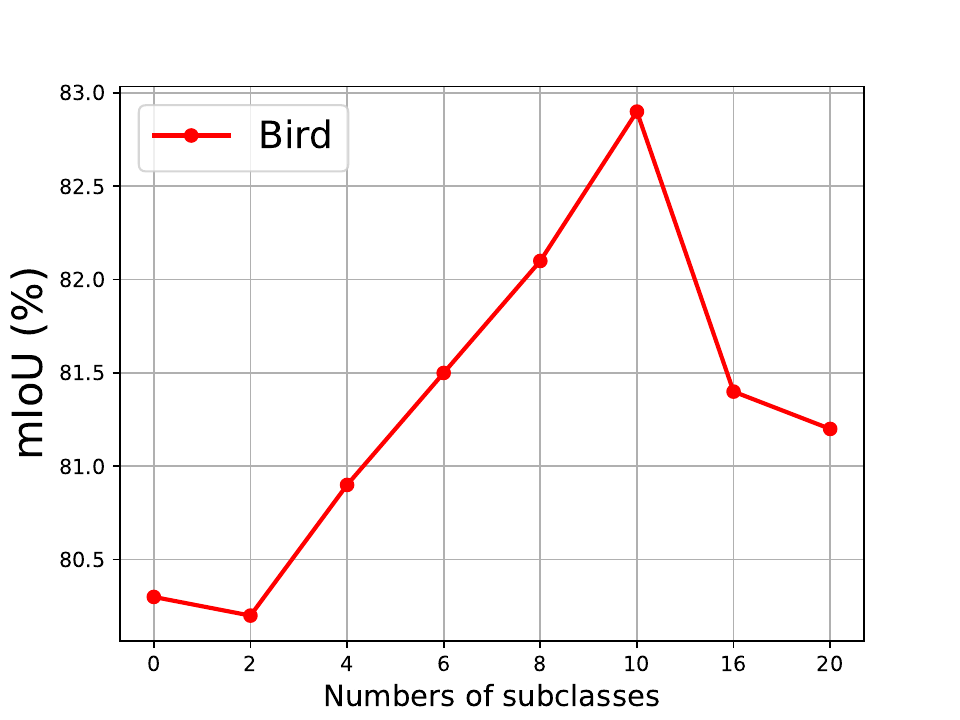}
            \caption*{Bird}
        \end{minipage} \\
    \end{tabular}

       \begin{tabular}{cccc}

        \begin{minipage}{0.22\textwidth}
            \includegraphics[width=\linewidth]{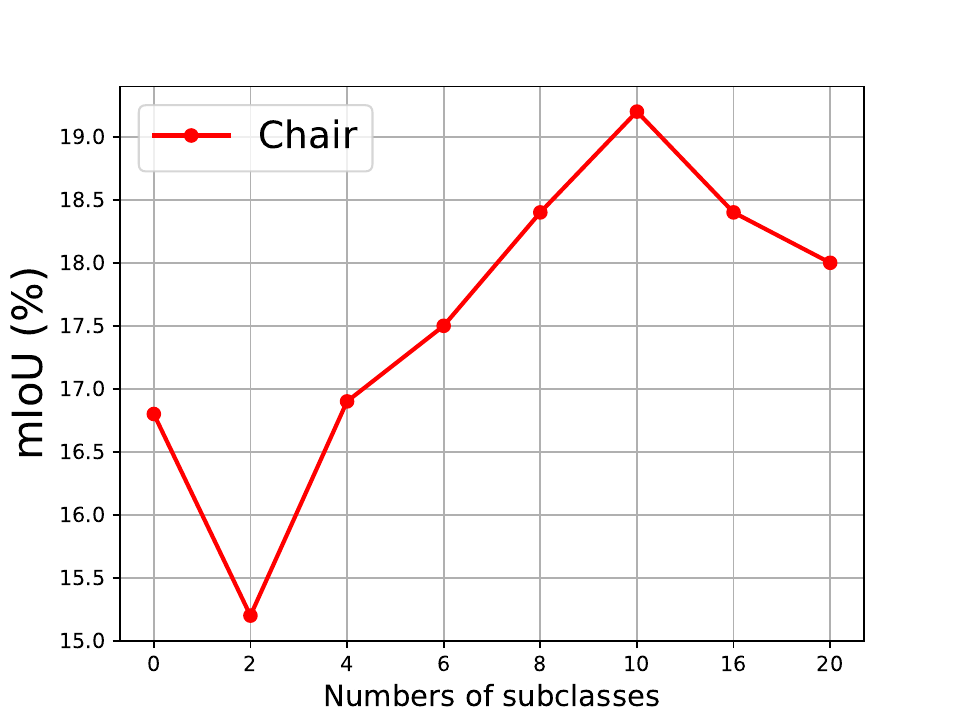}
            \caption*{Chair}
        \end{minipage} &
        \begin{minipage}{0.22\textwidth}
            \includegraphics[width=\linewidth]{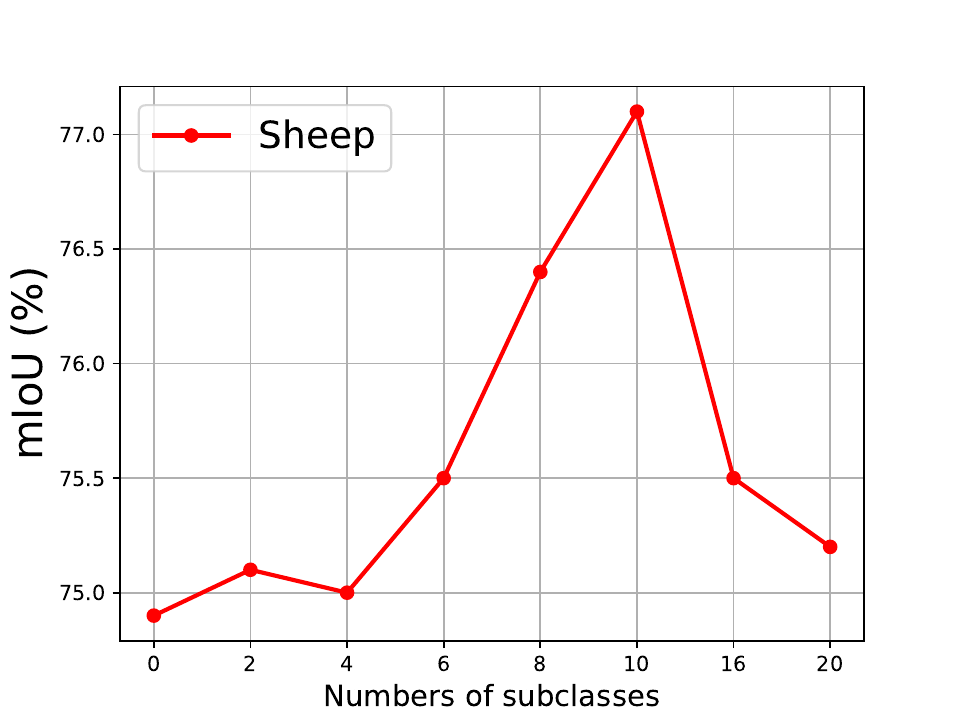}
            \caption*{Sheep}
        \end{minipage} &
        \begin{minipage}{0.22\textwidth}
            \includegraphics[width=\linewidth]{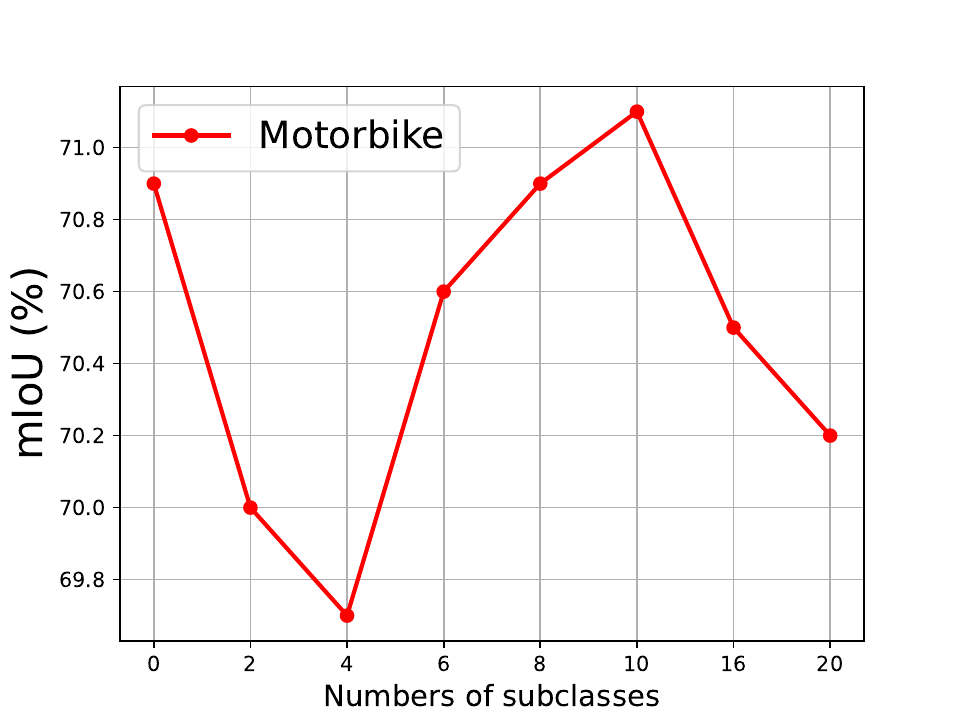}
            \caption*{Motorbike} 
            
        \end{minipage}
            &

                    \begin{minipage}{0.22\textwidth}
            \includegraphics[width=\linewidth]{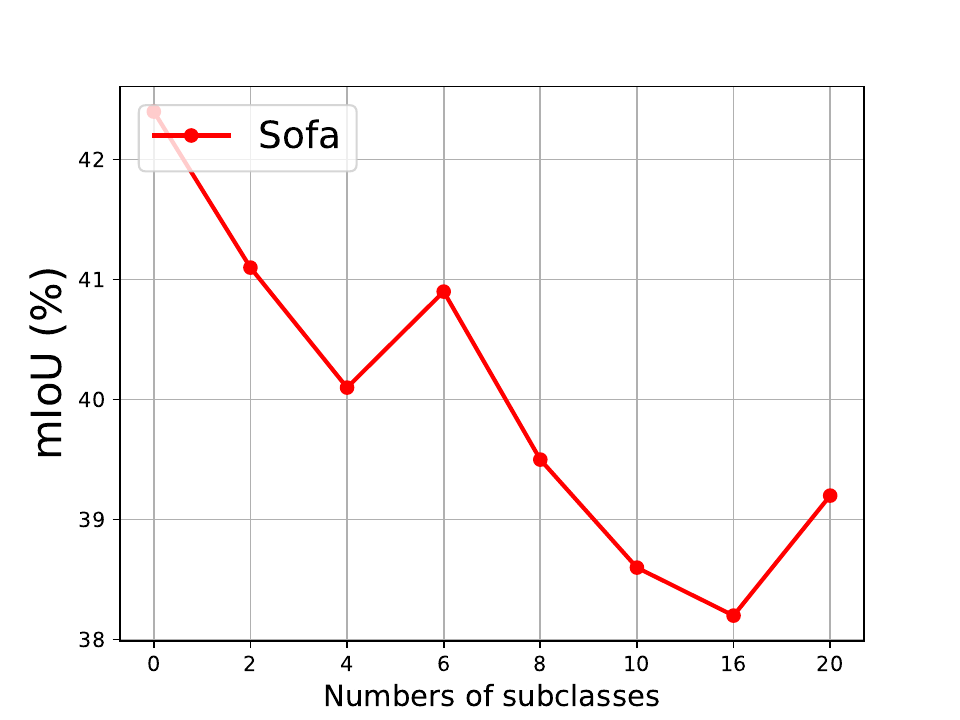}
            \caption*{Sofa}

        \end{minipage} \\
    \end{tabular}

    \caption{\textbf{Effect of the number of generated subclasses for each superclass.} An increase in the number of subclasses leads to a higher mean Intersection over Union (mIoU).}
    \label{fig:sub_Avg}

\end{figure*}

This section presents a comprehensive visual analysis through a series of images in Figure~\ref{fig:sub_Avg}, each corresponding to different superclasses like \textit{Person}, \textit{Boat}, \textit{Aeroplane}, \textit{Cow}, and many more. The layout is structured into multiple rows, with each row containing a set of images representing different superclasses.
Each image in the sequence is dedicated to a specific superclass and presumably showcases the effect of various subclasses within that category on the mIoU metric. 
For example, the first row might include images representing the \textit{Person}, \textit{Boat}, \textit{Aeroplane}, and \textit{Cow} superclasses, each illustrating how different subclasses within these categories impact the mIoU. Similarly, subsequent rows cover other superclasses like \textit{Table}, \textit{Horse}, \textit{Bottle}, \textit{Monitor}, and so forth, extending to categories like \textit{Car}, \textit{Bus}, \textit{Train}, \textit{Cat}, \textit{Dog}, \textit{Bicycle}, \textit{Plant}, \textit{Bird}, \textit{Chair}, \textit{Sheep}, \textit{Motorbike}, and \textit{Sofa}. These results are better to understand the relationship between subclass diversity and segmentation performance in various contexts. 

\FloatBarrier

\section{Additional Results}

\begin{figure*}[h]
    \centering
   \includegraphics[width=1.0\textwidth]{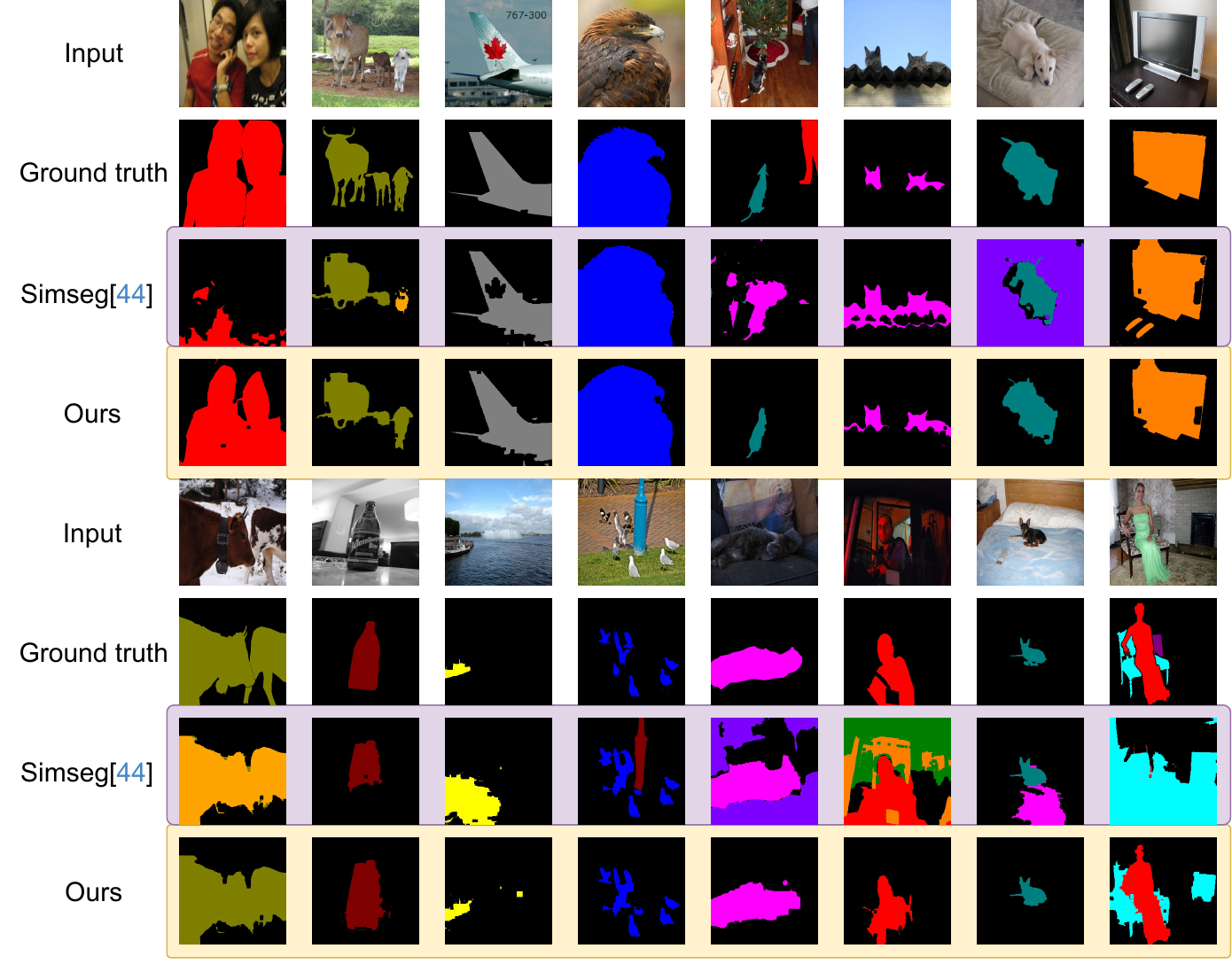}
\caption{\textbf{Segmentation Results with Our LLM-Supervision.} The use of subclass textual representations leads to more informative and precise segmentation outcomes compared to those achieved with superclass textual representations.}
\label{fig:success}
\vspace{-4mm}
\end{figure*}

This section includes further visual demonstrations of the segmentation results, as shown in Figure~\ref{fig:success}. When utilizing text-supervision based solely on superclasses, the segmentation typically focuses on the overarching traits of each category. However, our approach, which employs LLM supervision, excels in identifying more intricate details due to the inclusion of generated subclasses.
The enhancements observed in the segmentation outcomes are credited to the LLM-supervision methodology. This approach produces richer and more descriptive class representations, leading to noticeable improvements compared to relying solely on superclass text-supervision.

\FloatBarrier

\end{document}